\def\paperlanguage{}
\pdfoutput=1 % for arxiv

\documentclass[letterpaper, 10 pt, conference]{ieeeconf}  % Comment this line out if you need a4paper

\usepackage{bm}
\usepackage{cite}
\include{preamble}

\newcommand{\ctext}[1]{\raise0.2ex\hbox{\textcircled{\scriptsize{#1}}}}
\definecolor{bvqa}{rgb}{0.7,0.7,0}
\definecolor{mvqa}{rgb}{0.7,0,0}
\definecolor{itr}{rgb}{0.7,0,0.7}
\definecolor{vg}{rgb}{0,0,0.7}
\definecolor{dic}{rgb}{0,0.7,0.7}

\IEEEoverridecommandlockouts                              % This command is only needed if
\title{\LARGE \textbf
  {
    \switchlanguage%
    {%
      GeMuCo: Generalized Multisensory Correlational Model\\ for Body Schema Learning
    }%
    {%
      身体図式学習学習に向けた一般化多感覚相関モデル
    }%
  }
}

\author{Kento Kawaharazuka$^{1}$, Kei Okada$^{1}$, and Masayuki Inaba$^{1}$% <-this % stops a space
  \thanks{$^{1}$ The authors are with the Department of Mechano-Informatics, Graduate School of Information Science and Technology, The University of Tokyo, 7-3-1 Hongo, Bunkyo-ku, Tokyo, 113-8656, Japan.
    {\texttt\small [kawaharazuka, k-okada, inaba]@jsk.t.u-tokyo.ac.jp}
  }
}
\begin{document}

\maketitle
\thispagestyle{empty}
\pagestyle{empty}

%%%%%%%%%%%%%%%%%%%%%%%%%%%%%%%%%%%%%%%%%%%%%%%%%%%%%%%%%%%%%%%%%%%%%%%%%%%%%%%%
\begin{abstract}
  \switchlanguage%
  {%
    Humans can autonomously learn the relationship between sensation and motion in their own bodies, estimate and control their own body states, and move while continuously adapting to the current environment.
    On the other hand, current robots control their bodies by learning the network structure described by humans from their experiences, making certain assumptions on the relationship between sensors and actuators.
    In addition, the network model does not adapt to changes in the robot's body, the tools that are grasped, or the environment, and there is no unified theory, not only for control but also for state estimation, anomaly detection, simulation, and so on.
    In this study, we propose a Generalized Multisensory Correlational Model (GeMuCo), in which the robot itself acquires a body schema describing the correlation between sensors and actuators from its own experience, including model structures such as network input/output.
    The robot adapts to the current environment by updating this body schema model online, estimates and controls its body state, and even performs anomaly detection and simulation.
    We demonstrate the effectiveness of this method by applying it to tool-use considering changes in grasping state for an axis-driven robot, to joint-muscle mapping learning for a musculoskeletal robot, and to full-body tool manipulation for a low-rigidity plastic-made humanoid.
  }%
  {%
    人間は自律的に自身の身体の感覚と運動の関係を学習することで, 身体状態を推定・制御し, 常に現在環境に適応しながら動き続けることができる.
    一方, 現在のロボットは人間が記述したネットワーク構造を経験から学習し制御を行っており, 感覚と運動の関係性に一定の仮定を置いている.
    また, そのモデルはロボットの身体や把持した道具, 環境の変化に適応して変化せず, 制御だけでなく, 状態推定や異常検知, シミュレーション等に関する統一的な理論もない.
    そこで本研究では, 感覚と運動の相関関係を記述する身体図式を, ロボット自身が経験から, ネットワーク入出力等のモデル構造を含め獲得する一般化多感覚相関モデル(GeMuCo)を提案する.
    この身体図式モデルをオンライン更新することで現在環境に適応するとともに, その身体状態を推定・制御し, 異常検知やシミュレーション等まで行う.
    本手法を, 軸駆動型ロボットの把持状態を考慮した道具操作, 筋骨格型ロボットの関節-筋空間マッピング学習, 低剛性樹脂製ヒューマノイドの全身道具操作に適用することで, その有効性を示す.
  }%
\end{abstract}

\section{INTRODUCTION}\label{sec:introduction}
\begin{figure}[t]
  \centering
  \includegraphics[width=0.9\columnwidth]{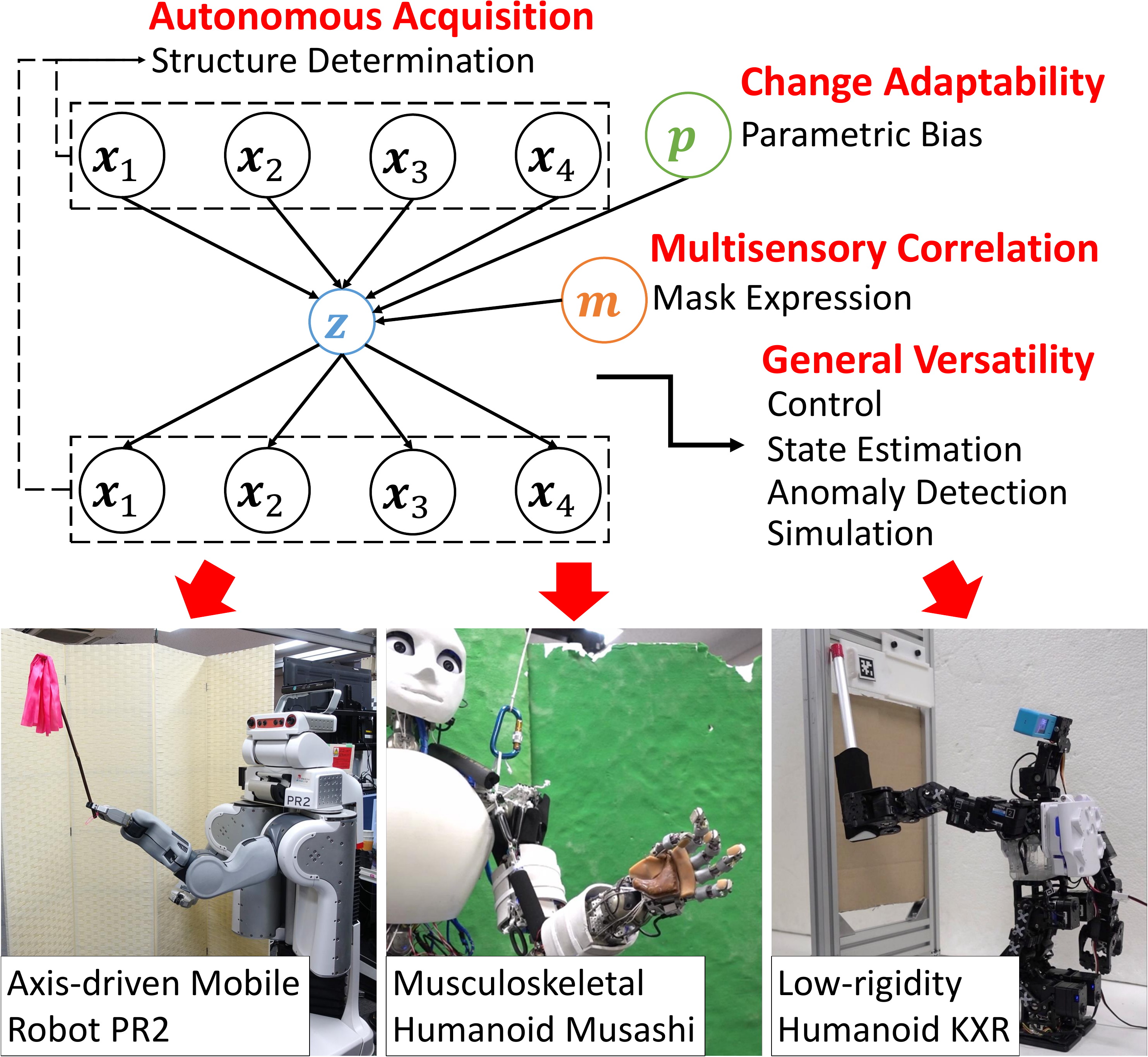}
  \caption{The concept of this study. Generalized Multisensory Correlational Model (GeMuCo) has four characteristics: Multisensory Correlation, General Versatility, Autonomous Acquisition, and Change Adaptability. This body schema model is used for motion control, state estimation, anomaly detection, and simulation of robots with various configurations.}
  \label{figure:concept}
  \vspace{-1.0ex}
\end{figure}

\switchlanguage%
{%
  Humans can continuously estimate and control their physical state by autonomously learning the relationship between various sensations and movements of their body, allowing them to adapt to the current environment and keep moving.
  A human can compensate for the lack of some their senses by using various other senses, just as a blindfolded human can roughly determine the position of their hand or a grasped tool from their own proprioception.
  Even in a complex body structure with joints, muscles, and tendons, the body gradually learns how to move by constantly learning the correlation between sensation and motion in an autonomous manner.
  Even if the body changes over time, or if the tools to be manipulated or the state of grasping changes, a human can detect and adapt to these changes, and can always estimate and control the state of the body appropriately.
  In this study, this human function is expressed in terms of a body schema, which represents the relationship between various sensors and actuators considering the body structure \cite{haggard2005bodyschema, hoffmann2010bodychema}.
  We have considered the following four requirements that a body schema should have in an intelligent robot system (\figref{figure:concept}).
  \begin{itemize}
    \item \textbf{Multisensory Correlation} - capable of expressing correlation among various sensors and actuators
    \item \textbf{General Versatility} - can be used to construct basic components from control to state estimation, anomaly detection, and simulation
    \item \textbf{Autonomous Acquisition} - enables acquisition of models, including network structures, through autonomous learning
    \item \textbf{Change Adaptability} - capable of coping with gradual changes in body schema by updating the model online
  \end{itemize}
  Note that \textbf{Multisensory Correlation} and \textbf{Change Adaptability} come from the fundamental features of body schema outlined in \cite{haggard2005bodyschema}, specifically termed as "Adaptable" and "Supramodel."
  Also, \textbf{General Versatility} comes from the perspective of applying the definition of body schema to robots, while \textbf{Autonomous Acquisition} comes from the context of the human process of acquiring the body schema.

  In the following, we discuss previous research related to these four requirements and the body schema itself, and our contributions.
}%
{%
  人間は自律的に自身の身体の多様な感覚と運動の関係を学習することで, 身体状態を推定・制御し, 常に現在環境に適応しながら動き続けることができる.
  目を隠しても自身の身体感覚から大体の手先の位置や道具の位置が分かるように, 一部のセンサ値が得られなくても, 他の多様な感覚からそれを補うことができる.
  関節と筋肉, 腱を持つ複雑な身体構造であっても, 自律的に感覚と運動の相関関係を常に学習することで, 徐々に身体の動かし方を分かっていく.
  身体が経年劣化によって変化したり, 操作する道具や把持状態が変化してもそれらを検知・適応し, 常に状態を推定し制御することができる.
  本研究ではこの人間の機能を, 人間における感覚と運動の関係を身体構造をもって表現した身体図式という言葉で表す\cite{head1911bodyschema, hoffmann2010bodychema}.
  知能ロボットシステムにおいて身体図式が持つべき要件は以下の4つであると考えた(\figref{figure:concept}).
  \begin{itemize}
    \item \textbf{多感覚相関性} - 多様な感覚間の相関を表現可能
    \item \textbf{汎用性} - 制御から状態推定, 異常検知, シミュレーション等のロボットの基本要素を構成可能
    \item \textbf{自律獲得性} - 自律的な学習によりネットワーク構造を含めたモデル獲得が可能
    \item \textbf{変化適応性} - モデルの逐次更新により逐次的なモデル変化に対処可能
  \end{itemize}
  なお, 多感覚相関性と変化適応性は\cite{haggard2005bodyschema}における身体図式の基礎的な特徴(``Adaptable''と``Supramodel'')から, 汎用性は身体図式の定義をロボットに応用するという観点から, 自律獲得性は身体図式を人間が獲得していくプロセスの観点から, これらを要件として定義している.

  空間知覚性は身体図式が身体部位を空間内の体積を持った物体としてその位置や配置を表現していること, モジュラー性は身体図式が身体部位ごとにモジュール化されていること, 動作更新性は身体図式が常に体の動きに従ってそれを認識し更新されること, 変化適応性は怪我や成長に伴って身体図式が変化すること, 多感覚性は身体図式が多様な感覚情報を統合していること, 時空間一貫性は身体図式が時間的かつ空間的に一貫した構造を提供すること, 他者間共通性は共通の身体図式がそれぞれの人間に共有されていることを表す.

  本研究では, 感覚入力や制御入力の間の関係性を表現し, 体をどう動かすと視界や接触, 音や温度がどう変化するといった相関を記述するモデルを身体図式と呼び, これを逐次的に学習することでロボットの環境適応能力を向上させることを考える.
  この身体図式を核として, ロボットの制御や状態推定, 異常検知やシミュレーションを行う.
  本研究の身体図式は心理学や神経科学における身体図式の一部の特徴を知能ロボットシステムの観点から抽出したものと言える.
  特に, 空間知覚性, 他者間共通性は取り扱わない.
  空間知覚性については有用な場面があるものの, 本研究の感覚入力や制御入力の相関表現には必要ないため扱っていない.
  動作更新性, 時空間一貫性, モジュラー性の特徴は本研究の身体図式も有している一方で, これらはロボットのモデル化としては当たり前であるため, 明示的には取り扱わない.
  よって, 多感覚性と変化適応性のみを明示的に考慮する.
  一方で, これまでの議論から, この身体図式は身体-道具-動作環境に関する相関複雑性と時間的変容に対応しなければならない.
  つまり, この身体図式のモデル化という観点では, その\textbf{モデル化困難性}と\textbf{逐次的モデル変化}に対応しなければならないということを意味する.
  また, 様々な身体部位や道具, タスク等を扱うことが可能な\textbf{一般性}にも対応する必要がある.
}%

\switchlanguage%
{%
  \subsection{Multisensory Correlation}
  \textbf{Multisensory Correlation} is not simply a matter of dealing with multiple sensors.
  It is necessary to express correlations among various sensors so that the model can cope with situations such as when a certain sensor is unavailable, or when a certain control input is not desired to be used.
  On the other hand, most of existing methods use proprioception, visual, tactile, and other sensors as network inputs at one time for end-to-end processing \cite{zhang2015reinforcement, lathuiliere2018audiovisual, zambelli2020multimodal}, and there is limited research directly addressing correlations of multiple modalities.

  \subsection{General Versatility}
  For \textbf{General Versatility}, previous learning methods have generally been structured to achieve one goal, e.g. motion control \cite{levine2018grasping}, recognition \cite{krizhevsky2012imagenet}, and anomaly detection \cite{park2018anomaly}.
  On the other hand, there are few examples of integrating these multiple tasks in a single network.
  If control, state estimation, anomaly detection, simulation, etc. can be handled in a unified manner using a general-purpose computation procedure based on a single model, the learning results can be uniformly reflected in these components, and manageability can be improved.

  \subsection{Autonomous Acquisition}
  For \textbf{Autonomous Acquisition}, it is important not to use manually constructed assumptions for each robot, no matter how complex the body structure is.
  Although various learning-based control methods have been developed for complex robots such as those with flexible links \cite{sum2017flexible}, these methods make many assumptions on the structure of the problem and cannot be used for musculoskeletal structures, flexible tools, flexible objects, etc. in a unified manner.
  While there have been many attempts using imitation learning \cite{wu2018flexible} and reinforcement learning \cite{hoof2015reinforcement}, there are limited methods that allow actual robots to learn from experience without human intervention.
  Also, autonomous acquisition should not be limited to the mere acquisition of a model through learning.
  For truly autonomous acquisition, it is necessary for a robot to acquire even the structure of the model itself autonomously.
  Although the most common network structure determination method is Neural Architecture Search (NAS) for deep neural networks \cite{zoph2017nas}, these are parameter searches of neural networks, and the input/output of the network is generally fixed.
  Note that some methods utilizing mutual information and self-organizing maps have also been developed \cite{bowden2005input, kobayashi2019automatic}.

  \subsection{Change Adaptability}
  For \textbf{Change Adaptability}, existing methods rarely incorporate information about gradual model changes directly into the models \cite{bongard2006resilient, cully2015adapt}.
  In reinforcement learning \cite{zhang2015reinforcement} and supervised learning \cite{levine2018grasping}, adaptive behavior is generated by collecting a large amount of data in various environments.
  In other words, if the purpose of the behavior is uniquely determined, adaptive behavior can be produced by collecting data in various environmental conditions.
  On the other hand, if the purpose of the behavior is not uniquely determined and the model is to be used for various control, state estimation, anomaly detection, etc., it is difficult to obtain the change adaptability by simply inputting a large amount of data into the network.

  \subsection{Body Schema in Robotics}
  Research on body schema learning in robotics has been conducted extensively \cite{hoffmann2010bodychema, hoffmann2022bodymodel}.
  The simplest example of body schema would be a generic robot model.
  Efforts to identify models with parameters such as link length, weight, and inertia have been numerous \cite{hollerbach2016identification}.
  However, these efforts are limited to easily modelable robots and cannot handle \textbf{Multisensory Correlation}.
  There have also been attempts at body schema learning in more challenging setups, such as when the link structure is unknown \cite{sturm2009bodyschema}, but these still make numerous assumptions about the robot's body structure.
  Recently, body schema learning methods using deep learning have been developed to overcome these limitations \cite{zambelli2020multimodal}, but the network structure is human-designed and lacks \textbf{Autonomous Acquisition} and \textbf{Change Adaptability}.
  Conversely, methods that heavily address \textbf{Change Adaptability} often lack \textbf{Multisensory Correlation} and \textbf{General Versatility} \cite{bongard2006resilient, cully2015adapt}.
  Furthermore, alongside the recent development of foundational models, models that learn manipulation strategies end-to-end based on diverse sensory and linguistic inputs have emerged \cite{driess2023palme}.
  However, due to these models being very large and learning policies directly from teaching data, they lack \textbf{General Versatility}, \textbf{Autonomous Acquisition}, and \textbf{Change Adaptability}.
}%
{%
  多感覚性は単純に多感覚を扱えば良いという話ではない.
  多数の感覚の相関関係を表現することで, あるセンサが見えない状態や, ある制御入力を使いたくない状態等にも対応できるようにする必要がある.
  一方で, 既存の手法は基本的に内部感覚や視覚, 触覚等を一度に入力してEnd-to-Endに用いる場合が多く\cite{zhang2015reinforcement, lathuiliere2018audiovisual, zambelli2020multimodal}, 相関関係を直接扱った研究は少ない.

  汎用性について, これまでの学習手法は基本的に, 制御\cite{levine2018grasping}や認識\cite{krizhevsky2012imagenet}, 異常検知\cite{park2018anomaly}等, 一つの目的を達成するために考えられてきた.
  一方で, これらを一つのネットワークで統合した例は少ない.
  制御や状態推定, 異常検知, シミュレーション等を一つのモデルをもとに汎用的な計算手順で統一的に扱うことができれば, 学習結果を一様にそれらコンポーネントに反映することができ, 管理性が向上する.

  自律獲得性について, どんなに複雑な身体構造に対してもロボットごとに手動で構築した仮定を用いないことが重要である.
  これまでリンク構造推定\cite{sturm2009bodyschema}, 柔軟物体操作\cite{yamakawa2010knotting}や柔軟アーム操作\cite{hofer2019iterative}に関する研究が行われてきたが, 問題の構造に多くの仮定を置いており, 統一的に筋骨格構造や柔軟道具, 柔軟物体等に利用できるようなものではない.
  模倣学習や強化学習の試みも多い一方で, 人間の介入なく, 実機でロボットが経験から学習していくことが可能な手法は非常に限られる.
  また, 自律獲得性は単に学習によりモデルを獲得するのではなく, そのモデルの構造自体までもロボットが自律的に獲得するべきである.
  最も一般的に行われているネットワーク構造決定は, ニューラルネットワークのパラメータ探索であり, 近年はNeural Architecture Searchの分野で議論が進んでいる\cite{zoph2017nas}.
  一方, これらはニューラルネットワークのパラメータ探索であり, 基本的にネットワークの入出力は固定されている.
  なお, 一部相互情報量や自己組織化マップを用いた手法も開発されている\cite{bowden2005input, kobayashi2019automatic}.

  変化適応性について, 逐次的なモデルの変化に関する情報を直接モデルに取り入れる手法は非常に少ない\cite{bongard2006resilient, cully2015adapt}.
  強化学習や教師あり学習は, 様々な環境において大量のデータを収集することで適応的な動作を生み出す.
  つまり, 動作の目的が一意に決まっている場合は様々な環境状態でデータを取ることで適応的な動作が可能となる.
  一方, 本研究のように動作の目的が一意に決まっておらず, そのモデルを使って様々な制御や状態推定, 異常検知等を行う汎用性を持たせる場合は, 単に大量のデータをネットワークに入力しても効果を得るのは難しい.

  ロボットにおける身体図式学習については, これまでも様々な研究が行われてきた\cite{hoffmann2010bodychema, hoffmann2022bodymodel}.
  身体図式の最も単純な例は一般的なロボットモデルだろう.
  リンクの長さや重さ, 慣性などのパラメータを持つモデルを同定する試みは古くから行われてきた\cite{hollerbach2016identification}.
  一方, これらはモデル化が容易なロボットのみへの適用に限られているうえ, 多感覚性を扱うことができない.
  リンク構造が分からないなど, より難しい設定における身体図式学習の試みもあるが\cite{sturm2009bodyschema}, ロボットの身体構造に多数の仮定を置いていることには変わりはない.
  これを打破する方法論として深層学習を利用した身体図式学習手法が開発されているが\cite{zambelli2020multimodal}, ネットワーク構造は人間が記述しており自律獲得性に欠け, 変化適応性にも問題がある
  逆に変化適応性を大きく扱った手法は, 汎用性や多感覚性に欠けてしまう\cite{bongard2006resilient, cully2015adapt}.
  また, 近年は基盤モデルの発展とともに, 多様な感覚と言語を用いてEnd-to-Endにマニピュレーション戦略を学習するモデルが登場している\cite{driess2023palme}.
  一方, 非常に巨大なモデルかつ教示データから直接ポリシーを学習しているため, これらは汎用性や自律獲得性, 変化適応性を失っている.
}%

\begin{figure*}[t]
  \centering
  \includegraphics[width=1.9\columnwidth]{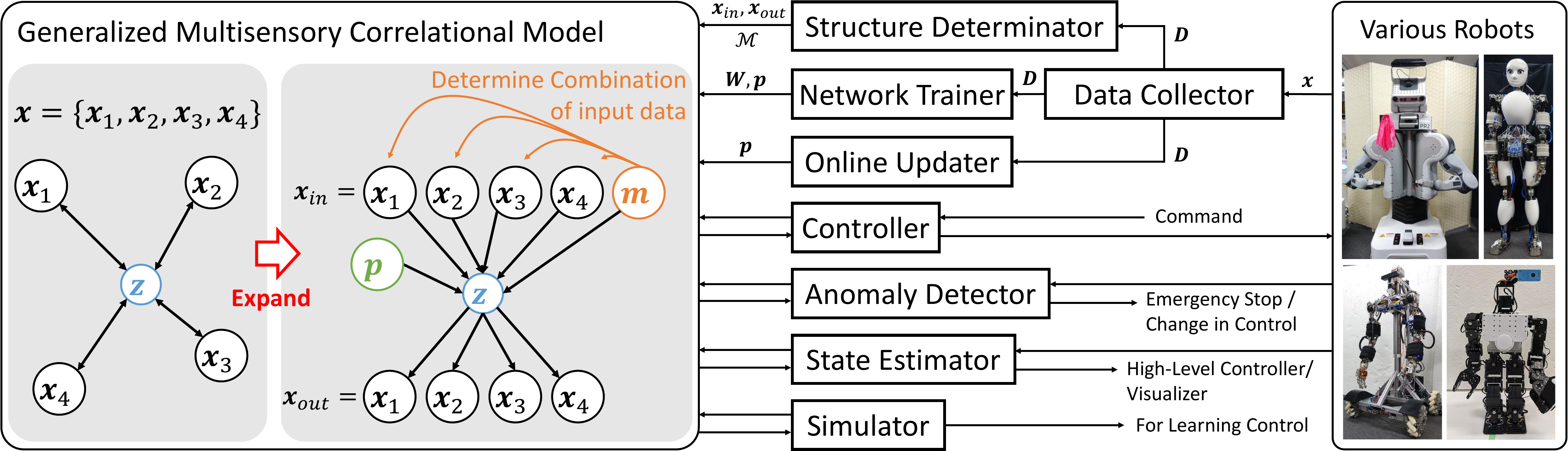}
  \caption{System overview of GeMuCo with various basic components of Data Collector, Structure Determinator, Network Trainer, Online Updater, Controller, Anomaly Detector, State Estimator, and Simulator.}
  \label{figure:system-overview}
\end{figure*}

\switchlanguage%
{%
  \subsection{Our Contributions}
  From the above discussion, a body schema learning method that satisfies all the four requirements does not exist.
  Therefore, the purpose of this study is to improve the robot's adaptive ability by modeling a body schema with these characteristics and having it learn autonomously.
  Drawing on our experiments which include tool-tip manipulation learning considering grasping state changes of axis-driven robots \cite{kawaharazuka2021tooluse}, joint-muscle  mapping learning for musculoskeletal humanoid robots \cite{kawaharazuka2020autoencoder}, and whole-body tool manipulation learning for low-rigidity humanoid robots \cite{kawaharazuka2024kxr}, we propose a theoretical framework that consolidates these, called the Generalized Multisensory Correlation Model (GeMuCo).
  In this model, when the values of sensors and actuators are collectively represented by the variable $\bm{x}$, the network structure is $(\bm{x}, \bm{m})\to\bm{x}$, using the mask variable $\bm{m}$ to represent the correlation between various sensors and actuators.
  This is a static body schema for static motions, i.e. motions for which there is a one-to-one correspondence between a certain control input and a sensation.
  Note that, in the case of dynamic motions, the network structure becomes a time-evolving dynamic body schema with different times at the input and output of the network \cite{kawaharazuka2023dpmpb}.
  In this study, we mainly discuss static body schema.
  Our contributions of GeMuCo are as follows:
  \begin{itemize}
    \item Multisensory correlation modeling by mask expression
    \item Versatile realization of control, state estimation, simulation, and anomaly detection by a single network
    \item Automatic acquisition of model structure including model input/output and their correlation
    \item Change adaptability by a mechanism of parametric bias
  \end{itemize}
  We hope that this study will contribute to the development of robots with autonomous learning capabilities similar to humans.
}%
{%
  これまで述べた全ての要件を満たすようなを身体図式学習手法は存在しない.
  よって, 本研究ではこれらの特徴を持った身体図式をモデル化し, 自律的に学習させ, ロボットの適応能力を向上させることを行う.
  これまで行ってきた, 軸駆動型ロボットの把持状態変化を考慮した道具先端操作学習\cite{kawaharazuka2021tooluse}, 筋骨格ヒューマノイドの関節-筋空間マッピング学習\cite{kawaharazuka2020autoencoder}, 低剛性樹脂製ヒューマノイドの全身道具操作学習を題材に, これらをまとめる理論体系である, 一般化多感覚相関モデル(GeMuCo)を提案する.
  これは感覚入力と制御入力の値を一緒くたに変数$\bm{x}$としたとき, その相関関係を表現するマスク変数$\bm{m}$を用いて, $(\bm{x}, \bm{m})\to\bm{x}$というネットワーク構造となる.
  これは静的な動作, つまりある制御入力と感覚が一対一対応するような動作に対する, 静的身体図式である.
  一方, 動的な動作の場合には, ネットワークの入出力で時刻が異なる, 時間発展する動的身体図式となる\cite{kawaharazuka2023dpmpb}.
  本研究では, 静的な身体図式学習を主に議論する.
  本研究の貢献は以下の通りである.
  \begin{itemize}
    \item マスク表現による多様なセンサ・アクチュエータ間の相関関係のモデル化
    \item 単一モデルによる制御・状態推定・シミュレーション・異常検知の汎用的実現
    \item モデル入出力とその相関関係を含むモデル構造の自動獲得
    \item Parametric Biasによる変化適応性
  \end{itemize}
  本研究が, 人間同様の自律学習機能の備わったロボットへの発展に寄与することを期待する.
}%

\section{GeMuCo: Generalized Multisensory Correlational Model} \label{sec:proposed}
\switchlanguage%
{%
  The overall system of Generalized Multisensory Correlational Model (GeMuCo) is shown in \figref{figure:system-overview}.
  First, various sensory and control input data are collected (Data Collector), and GeMuCo is trained based on these data (Network Trainer).
  In this process, the input/output of the network and the feasible mask set can be automatically determined from the data (Structure Determinator).
  During operation, GeMuCo always collects sensory and control input data and continuously updates part or all of it based on these data (Online Updater).
  Given a target state, GeMuCo calculates the control input to realize the state and commands to the robot (Controller).
  Based on some sensors and actuators, the current latent state of the robot is calculated, and sensor values that cannot be directly obtained are estimated (State Estimator).
  Anomaly detection is performed based on the prediction errors of the sensor values (Anomaly Detector).
  GeMuCo can also be used to simulate the actual robot behavior based on the control input (Simulator).
}%
{%
  本研究で提案する身体図式モデルであるGeneralized Multisensory Correlational Model (GeMuCo)を含む全体システムを\figref{figure:system-overview}に示す.
  まず, ロボット実機から多様な感覚と制御入力のデータを収集し(Data Collector), これをもとにGeMuCoを学習させる(Network Trainer).
  この際, ネットワークの入出力や実行可能なマスク集合をデータから自動決定することができる(Structure Determinator).
  実際の運用時は, 常に感覚と制御入力のデータを取得し, これをもとにGeMuCoの一部または全部を更新し続けることを行う(Online Updater).
  ある指令状態が与えられた場合, GeMuCoからこれを実現する制御入力を計算し, 実機に指令する(Controller).
  一部の感覚と制御入力から, 現在のロボットの潜在状態を計算し, 直接得られない感覚を状態推定(State Estimator), また, 感覚の予測誤差から異常検知を行う(Anomaly Detector).
  GeMuCoを使い制御入力に基づく実機動作のシミュレーションを行うことも可能である(Simulator).
}%

% \begin{figure}[t]
%   \centering
%   \includegraphics[width=0.95\columnwidth]{figs/network-structure}
%   \caption{The network structure of GeMuCo.}
%   \label{figure:network-structure}
% \end{figure}

\begin{figure}[t]
  \centering
  \includegraphics[width=1.0\columnwidth]{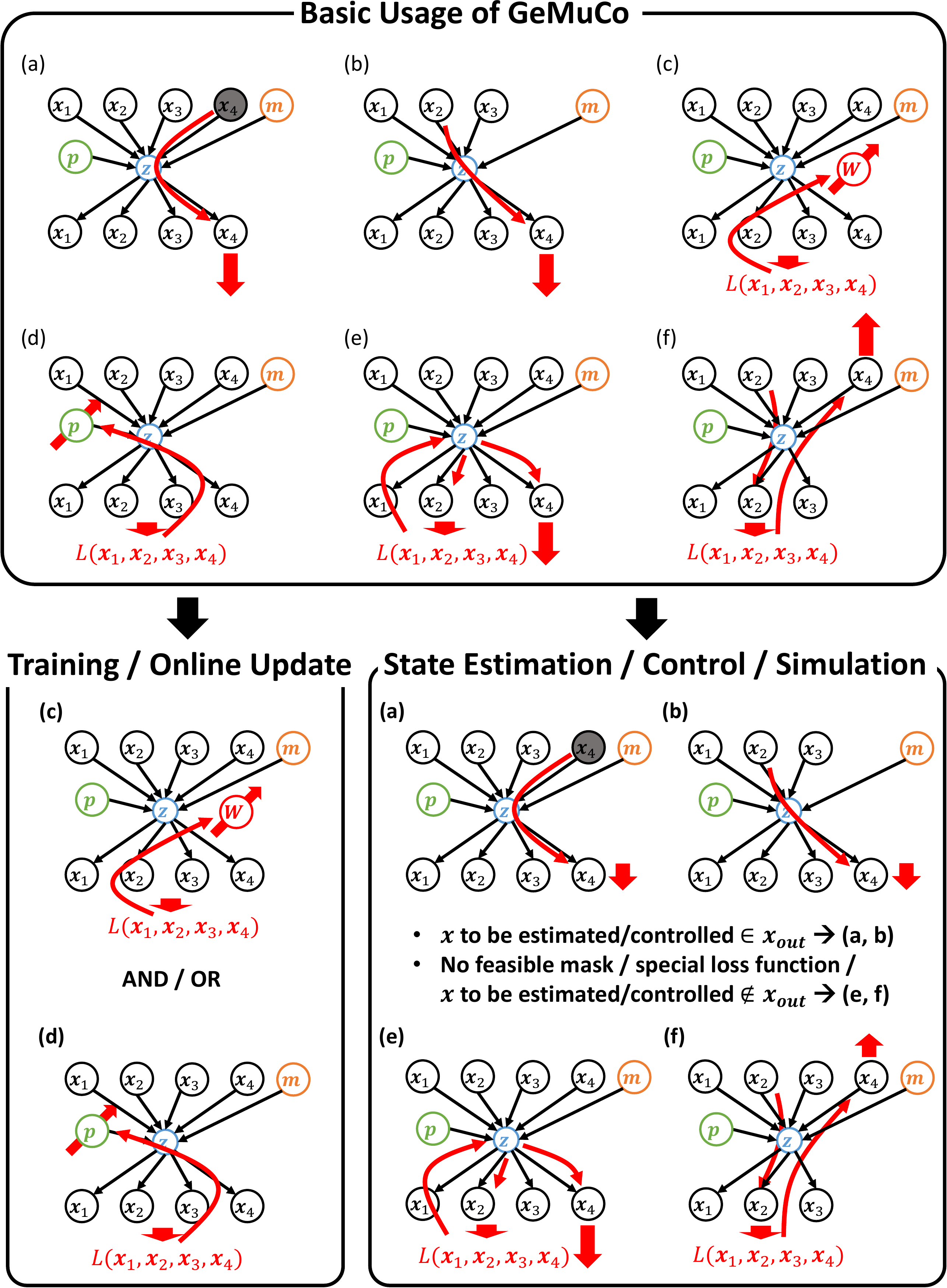}
  \caption{The six basic usages of GeMuCo and their application to the training and online update of GeMuCo and to the state estimation, control, and simulation using GeMuCo. The six usages include the types of forward and backward propagations for network input/output, and (a)--(f) can be used in combination with each other. (a) estimates data that is currently unavailable. (b) is almost the same method as (a), but it is used when the output data to be inferred is not available in the input. (c) updates the network weight $\bm{W}$. (d) updates parametric bias $\bm{p}$. (e) estimates $\bm{x}_{out}$ that optimizes a certain loss function by iterating forward and backward propagations. (f) is an iterative calculation similar to (e), but it corresponds to the case where the desired value is not available on the output side.}
  \label{figure:basic-usage}
\end{figure}

\subsection{Network Structure of GeMuCo} \label{subsec:network-structure}
\switchlanguage%
{%
  The network structure of GeMuCo is shown in the left figure of \figref{figure:system-overview}.
  First, we consider that there is a latent state $\bm{z}$ that can represent the current sensory and control inputs $\bm{x}$.
  This means that when there exist, e.g. four kinds of sensory and control inputs $\bm{x}=\bm{x}_{\{1, 2, 3, 4\}}$, all values of $\bm{x}_{\{1, 2, 3, 4\}}$ can be inferred by this $\bm{z}$.
  Moreover, this $\bm{z}$ can be inferred using some or all of $\bm{x}_{\{1, 2, 3, 4\}}$.
  This means that each of the sensory and control inputs are correlated with each other via $\bm{z}$.
  On the other hand, these relationships are difficult to handle as they are to construct a practical data structure of the model.
  Therefore, we expand this model and consider a model where the network inputs are $\bm{x}_{\{1, 2, 3, 4\}}$ and the mask variable $\bm{m}$, the middle layer is the latent state $\bm{z}$, and the network output is $\bm{x}_{\{1, 2, 3, 4\}}$.
  Here, we call the network from the input to the middle layer Encoder, and the network from the middle layer to the output Decoder.
  Let $\bm{h}$ denote the function of the entire model, $\bm{h}_{enc}$ the function of the Encoder, and $\bm{h}_{dec}$ the function of the Decoder.
  Note that $\bm{x}$ is assumed to be normalized using all the data obtained.

  \subsubsection{\textbf{Mask Variable}}
  $\bm{m}$ ($ \in {\{0, 1\}}^{N_{sensor}}$ where $N_{sensor}$ is the number of sensors and actuators) is a variable that masks the input $\bm{x}_{\{1, 2, 3, 4\}}$ .
  If $i$-th line of $\bm{m}$ is $0$, we set $\bm{x}_{i}=\bm{0}$ and mask it completely.
  On the other hand, if $\bm{x}_{i}$ is $1$, the current value $\bm{x}_{i}$ is used as network input.
  In other words, some of the inputs are masked by $\bm{m}$, and $\bm{z}$ is computed from a limited number of network inputs.
  This makes it possible to infer the masked values and to use them for state estimation and anomaly detection.
  Of course, not all $\bm{m}$ is acceptable, and it is necessary to maintain a set of feasible masks $\mathcal{M}$.
  Note that the input and output of the network need not be the same.
  In this study, we represent the values used for the network input as $\bm{x}_{in}$ and the values used for the network output as $\bm{x}_{out}$, and all the values used for the input/output of the network are expressed as $\bm{x}$.

  \subsubsection{\textbf{Parametric Bias}}
  As a characteristic structure, $\bm{p}$, parametric bias (PB) \cite{tani2002parametric}, is given as network input.
  This is a mechanism that has been mainly used in imitation learning and has been utilized in cognitive robotics research for the purpose of extracting multiple attractor dynamics from the obtained experience.
  The extraction of object dynamics from multimodal sensations \cite{ogata2005rnnpb} and the extraction of changes in hand-eye dynamics due to tool grasping \cite{nishide2009toolbody} are being conducted.
  On the other hand, we do not directly use the parametric bias in the context of imitation learning in this study.
  We embed information on changes in the body, tools, and environment into this parametric bias, and update it according to the current state to adapt to the environment.

  From the above, $\bm{h}$, $\bm{h}_{enc}$, and $\bm{h}_{dec}$ can be expressed as follows.
  \begin{align}
    \bm{z} &= \bm{h}_{enc}(\bm{x}_{in}, \bm{m}, \bm{p})\\
    \bm{x}_{out} &= \bm{h}_{dec}(\bm{z})\\
    \bm{x}_{out} &= \bm{h}(\bm{x}_{in}, \bm{m}, \bm{p})
  \end{align}

  % In this study, since we deal with static body schema, the network input and output are the values at the same time step.
  % On the other hand, when dealing with dynamic body schema, the network consists of $\bm{x}_{t}$ for input and $\bm{x}_{t+1}$ for output ($t$ represents the current time step).
  % In other words, the input of GeMuCo is $\bm{x}_{in, t}$, $\bm{m}$, and $\bm{z}_{t}$, the middle layer is $\bm{z}_{t+1}$, and the output is $\bm{x}_{out, t+1}$.
  % The dynamic body schema converges to either the state equation-type $(\bm{s}_{t}, \bm{u}_{t})\to\bm{s}_{t+1}$, or the imitation learning-type $(\bm{s}_{t}, \bm{u}_{t})\to(\bm{s}_{t+1}, \bm{u}_{t+1})$, when the sensor value in $\bm{x}$ is $\bm{s}$ and the control input is $\bm{u}$.
  % These are not handled explicitly in this study.
}%
{%
  GeMuCoのネットワーク構造を\figref{figure:system-overview}の左図に示す.
  まず, ロボットには現在の感覚や制御入力$\bm{x}$を表現可能な潜在状態$\bm{z}$が存在すると考える.
  つまり, 例えば4種類の感覚や制御入力$\bm{x}=\bm{x}_{\{1, 2, 3, 4\}}$が存在するとき, この$\bm{z}$によって$\bm{x}_{\{1, 2, 3, 4\}}$の全ての値が推論できるということを意味する.
  また, この$\bm{z}$は$\bm{x}_{\{1, 2, 3, 4\}}$のうちの一部, または全部を使って推論することができる.
  これは, $\bm{z}$を介してそれぞれの感覚や制御入力が互いに相関を持っていることを意味する.
  一方, 実際にこのモデルを作成することを考えた時, このままではデータ構造としての構築が難しい.
  そのため, このモデルを展開し, 入力が$\bm{x}_{\{1, 2, 3, 4\}}$とマスク変数$\bm{m}$, 中間層が潜在表現$\bm{z}$, 出力が$\bm{x}_{\{1, 2, 3, 4\}}$であるモデルを考える.
  ここで, 入力から中間層を求めるまでのネットワークをEncoder, 中間層から出力を求めるまでのネットワークをDecoderを呼ぶこととする.
  全体のモデルの関数を$\bm{h}$, Encoderの関数を$\bm{h}_{enc}$, Decoderの関数を$\bm{h}_{dec}$とする.
  なお, 以降$\bm{x}$は得られた全てのデータを使い正規化されているものとする.

  $\bm{m}$は入力である$\bm{x}_{\{1, 2, 3, 4\}}$をマスクする役目を持ち, $\bm{m}$は$\bm{x}$が$N_{sensor}$種類ある場合に, $\bm{m} \in {\{0, 1\}}^{N_{sensor}}$となる変数である.
  $\bm{m}$の$i$行目が0である場合, $\bm{x}_{i}=\bm{0}$とし, 完全にマスクする.
  一方, 1である場合は現在の値$\bm{x}_{i}$をそのまま入力として用いる.
  つまり, $\bm{m}$によって一部の入力をマスクし, 限られた入力から$\bm{z}$を計算することになる.
  これにより, 限られた感覚や制御入力から$\bm{z}$を計算して, マスクされた値を推論し, 状態推定や異常検知に利用することができるようになる.
  もちろん, どのような$\bm{m}$でも良いわけではなく, 実行可能な$\bm{m}$の集合$\mathcal{M}$を保持しておく必要がある.
  なお, ネットワークの入力と出力は同じである必要はない.
  本研究では, ネットワークの入力に使われる値を$\bm{x}_{in}$, 出力に使われる値を$\bm{x}_{out}$として表現し, ネットワークの入出力に用いる全ての値を並べたものを$\bm{x}$として表現する.

  また, 特徴的な構造として, $\bm{p}$のParametric Bias (PB) \cite{tani2002parametric}が入力に与えられている.
  これは, 主に模倣学習において利用されていた構造であり, 得られたデータから複数のアトラクターダイナミクスを抽出する目的で, 認知ロボティクス研究に使われてきた.
  マルチモーダルな感覚からの物体ダイナミクスの抽出\cite{ogata2005rnnpb}, 道具把持による手先ダイナミクスの変化の抽出\cite{nishide2009toolbody}等が行われている.
  一方, 本研究では模倣学習の文脈で直接用いることはない.
  Parametric Biasに身体や道具, 環境の変化の情報を埋め込み, これを現在の状態に応じて更新することで環境に適応していく.

  これらから, $\bm{h}$, $\bm{h}_{enc}$, $\bm{h}_{dec}$を式で表すと以下のようになる.
  \begin{align}
    \bm{z} &= \bm{h}_{enc}(\bm{x}_{in}, \bm{m}, \bm{p})\\
    \bm{x}_{out} &= \bm{h}_{dec}(\bm{z})\\
    \bm{x}_{out} &= \bm{h}(\bm{x}_{in}, \bm{m}, \bm{p})
  \end{align}

  % 本研究では静的な身体図式を扱うため, 入力と出力が同時刻の値となっている.
  % 一方, 動的な身体図式を扱う場合は, 入力を$\bm{x}_{t}$, 出力を$\bm{x}_{t+1}$とするネットワークとなる($t$は現在のタイムステップを表す).
  % つまり, GeMuCoの入力は$\bm{x}_{in, t}$, $\bm{m}$, $\bm{z}_{t}$, 中間層は$\bm{z}_{t+1}$, 出力は$\bm{x}_{out, t+1}$となる.
  % 動的身体図式は, $\bm{x}$に含まれるセンサ値を$\bm{s}$,制御入力を$\bm{u}$とすると, $(\bm{s}_{t}, \bm{u}_{t})\to\bm{s}_{t+1}$の状態方程式型か, $(\bm{s}_{t}, \bm{u}_{t})\to(\bm{s}_{t+1}, \bm{u}_{t+1})$の模倣学習型に収束する.
  % 本研究ではこれらは陽に扱わない.
}%

\subsection{Basic Usage of GeMuCo} \label{subsec:basic-usage}
\switchlanguage%
{%
  The six basic usages (a)--(f) of GeMuCo and their application to the training and online update of GeMuCo and to the state estimation, control, and simulation using GeMuCo are shown in \figref{figure:basic-usage}.
  % These are the types of forward and backward propagations for network input/output, and (a)--(f) can be used in combination with each other.
  (a) and (b) relate to simple forward propagation, while (c)--(f) represent updates of values through repeated forward and backward propagations.
  Specifically, (c)--(f) comprehensively cover backpropagation with respect to network weight $\bm{W}$, parametric bias $\bm{p}$, latent space $\bm{z}$, and network input $\bm{x}_{in}$.
  Note that these are mere usages regarding the inference and updating of values, and when actually used in a robot, they should be employed in the form of \secref{subsec:online-updater} -- \secref{subsec:anomaly-detection}.

  (a) is a method for estimating data that is currently unavailable.
  For example, if $\bm{x}_{4}$ is not available, $\bm{x}_{4}$ is inferred from $\bm{x}_{\{1, 2, 3\}}$ by setting $\bm{m}$ to $\begin{pmatrix}1&1&1&0\end{pmatrix}^{T}$.
  In other words, the method masks the data that cannot be obtained and infers the data from the remaining data.

  (b) is almost the same method as (a), but it is used when the output data to be inferred is not available in the input.
  For example, if $\bm{x}_{4}$ needs to be obtained, it is inferred from $\bm{x}_{\{1, 2, 3\}}$ by setting $\bm{m}$ to $\begin{pmatrix}1&1&1\end{pmatrix}^{T}$.
  In other words, the method is the same as that of a normal neural network which infers data not in the input $\bm{x}_{in}$.

  (c) corresponds to the adjustment of the network weight $\bm{W}$.
  A loss function $L$ is defined with respect to the output, and the weight $\bm{W}$ is updated from $\partial{L}/\partial{\bm{W}}$ as in general learning.

  (d) corresponds to the adjustment of parametric bias $\bm{p}$.
  We define a loss function $L$ with respect to the output and update $\bm{p}$ from $\partial{L}/\partial{\bm{p}}$.
  While updating the weight $\bm{W}$ changes the structure of the entire network, updating the parametric bias $\bm{p}$ changes a part of the relationship and dynamics while preserving the overall structure of the network.

  (e) is equivalent to computing $\bm{x}_{out}$ that optimizes a certain loss function by iterating forward and backward propagations.
  First, $\bm{z}$ is calculated by (a), (b), etc.
  Here, for example, if $\bm{x}_{4}$ satisfying certain conditions needs to be computed, $\bm{x}_{\{1, 2, 3, 4\}}$ is inferred from $\bm{z}$, the loss function is defined, and $\bm{z}$ is updated from $\partial{L}/\partial{\bm{z}}$.
  By repeating this inference and update, $\bm{x}_{4}$ with minimum loss function can be computed.
  In other words, the method minimizes the loss function by iterative inference and update on the decoder side.

  (f) is an iterative calculation similar to (e), but it corresponds to the case where the desired value is not available on the output side.
  For example, if $\bm{x}_{4}$ to be obtained is not in the output, the network output is inferred from $\bm{x}_{\{1, 2, 3, 4\}}$, the loss function is defined, and $\bm{x}_{4}$ is updated from $\partial{L}/\partial{\bm{x}_{4}}$.
  By repeating this inference and update, $\bm{x}_{4}$ can be computed such that the loss function is minimized.
  In other words, the method minimizes the loss function by iterative inference and update in the entire network, including Encoder and Decoder.
}%
{%
  GeMuCoの基礎的な6つの利用方法(a)--(f)を\figref{figure:basic-usage}に示す.
  これはネットワーク入出力に対する順伝播と誤差逆伝播の種類を表現しており, この(a)--(f)は互いに組み合わせて利用することができる.
  (a)と(b)は単純な順伝播に関して, (c)--(f)は順伝播と誤差逆伝播の繰り返しによる更新を表す.
  特に(e)--(f)は, 更新可能なネットワーク重み$\bm{W}$, parametric bias $\bm{p}$, 潜在空間$\bm{z}$, ネットワーク入力$\bm{x}_{in}$のそれぞれに関する誤差逆伝播を網羅している.
  なお, これらは単なる値の推定と更新に関する利用方法であり, 実際にロボットに使用する際は\secref{subsec:online-updater}--\secref{subsec:anomaly-detection}の形で利用する.

  (a)は, 現在得ることができないデータを推定する方法である.
  例えば$\bm{x}_{4}$が得られなかった場合, $\bm{m}$を$\begin{pmatrix}1&1&1&0\end{pmatrix}^{T}$に設定することで, $\bm{x}_{\{1, 2, 3\}}$から$\bm{x}_{4}$を推論する.
  つまり, 得ることができないデータをマスクして残りのデータからそのデータを推論する方法である.

  (b)は(a)とほとんど同じであるが, 推論したい出力のデータが入力に無い場合に利用される.
  例えば$\bm{x}_{4}$を得たい場合, $\bm{m}$を$\begin{pmatrix}1&1&1\end{pmatrix}^{T}$に設定することで, $\bm{x}_{\{1, 2, 3\}}$から$\bm{x}_{4}$を推論する.
    つまり, 入力側$\bm{x}_{in}$にはないデータを推論する, 通常のニューラルネットワークと同じ形と言える.

  (c)はネットワークの重み$\bm{W}$の調節に相当する.
  出力に関して損失関数$L$を定義し, $\partial{L}/\partial{\bm{W}}$から一般的な学習と同様に重み$\bm{W}$を更新する.

  (d)はParametric Bias $\bm{p}$の調節に相当する.
  出力に関して損失関数$L$を定義し, $\partial{L}/\partial{\bm{p}}$から$\bm{p}$を更新する.
  重み$\bm{W}$の更新はネットワーク全体の構造を変化させてしまうのに対して, Parametric Bias $\bm{p}$の更新はネットワークの全体的な構造は保ったまま, 一部の関係性・ダイナミクスを変更する.

  (e)は順伝播と誤差逆伝播の繰り返しにより, ある損失関数を最適化する$\bm{x}_{out}$を計算することに相当する.
  まずは(a)や(b)等により$\bm{z}$を計算する.
  ここで, 例えばある条件を満たす$\bm{x}_{4}$を計算したい場合, $\bm{z}$から$\bm{x}_{\{1, 2, 3, 4\}}$を推論し, これらに対して損失関数を定義, $\partial{L}/\partial{\bm{z}}$から$\bm{z}$を更新する.
  この推論と更新を繰り返すことで損失関数を最小化するような$\bm{x}_{4}$を計算することができる.
  つまり, Decoder側における推論と更新の繰り返しによる損失関数最小化を意味する.

  (f)は(e)と同様の繰り返し計算であるが, 求めたい値が出力側にない場合に相当する.
  例えば求めるべき$\bm{x}_{4}$が出力にない場合, $\bm{x}_{\{1, 2, 3, 4\}}$から出力を推論し, これらに対して損失関数を定義, $\partial{L}/\partial{\bm{x}_{4}}$から$\bm{x}_{4}$を更新する.
  この推論と更新を繰り返すことで損失関数を最小化するような$\bm{x}_{4}$を計算することができる.
  つまり, EncoderとDecoderを含むネットワーク全体における推論と更新の繰り返しによる損失関数最小化を意味する.
}%

\subsection{Data Collection for GeMuCo} \label{subsec:data-collection}
\switchlanguage%
{%
  In order to train GeMuCo, the necessary data of $\bm{x}$ needs to be collected.
  There are two main methods: random action and human teaching.
  In random action, $\bm{x}$ is obtained from random control inputs.
  % Note that although random, the operation is performed within certain constraints, such as the maximum and minimum values of the control input.
  It is also possible to consider a mapping from some random number to the control input, and use this mapping to operate the robot while applying constraints.
  In human teaching, a human directly decides the motion commands by using VR devices, sensor gloves, GUI applications, and so on.
  Data can be collected more efficiently for tasks that are difficult to perform by random action.

  It is not necessary to collect data for all $\bm{x}$ at all times.
  For example, it is acceptable if the robot vision is occasionally blocked, or if there are long intervals between some of the data collections.
  It is also acceptable to install a special sensor only when collecting data for training purposes.
  Also, $\bm{x}$ to be collected is not necessarily limited to the information directly obtained from the robot's sensors.
  It is possible to process the values of existing sensors before inputting them to the network, e.g. object recognition results obtained from image information or sound information related to specific frequencies, etc.
}%
{%
  GeMuCoの学習には, それに必要な$\bm{x}$のデータを収集する必要がある.
  この方法には主に, ランダム動作と教示の2つが存在する.
  まず, ランダム動作は制御入力をランダムに与え, このときの$\bm{x}$を取得する.
  % なお, ランダムと言っても, 制御入力の最大値や最小値等の制約を加えたうえで動作を行う.
  また, あるランダムな数値から制御入力への写像を考え, これを使って制約を加えながら動作させることも可能である.
  次に, 教示はVRデバイスやセンサグローブ, GUIアプリケーション等を使って, 人間が直接動作指令を決めることを言う.
  ランダムでは難しいタスクに対しては, より効率的にデータを収集することができる.

  収集する$\bm{x}$は, 常に全てのデータが集まる必要はない.
  たまに視界が遮られたり, 一部のデータの得られる間隔が長かったりしても良い.
  また, 実際にロボット運用する際には得られないセンサ入力を, 学習用のデータ収集時のみ特別なセンサを取り付けて取得するのも良い.
  また, 集める$\bm{x}$はロボットのセンサから直接得られる情報だけとは限らない.
  例えば画像情報から得られた物体認識結果や特定の周波数に関する音情報など, 存在するセンサの値を処理したうえで入力する場合も考えられる.
}%

\subsection{Training of GeMuCo} \label{subsec:network-trainer}
\switchlanguage%
{%
  Although the training of GeMuCo described in this section is executed after the automatic determination of the network structure, it is explained first since it is necessary for the automatic structure determination as well.

  When data $D$ of $\bm{x}$ is obtained, the output is usually inferred by taking $\bm{x}_{in}$ as input and training GeMuCo so that it becomes close to $\bm{x}_{out}$ using the mean squared error as the loss function.
  Here, it is necessary to include a mask variable $\bm{m}$ in the training for GeMuCo.
  First, we prepare a set of feasible masks $\mathcal{M}$.
  Then, for each $\bm{x}_{in}$, we use each $\bm{m}$ in $\mathcal{M}$ to mask a part of the corresponding $\bm{x}_{in}$ and create $\bm{x}^{masked}_{in}$.
  By inputting $\bm{x}^{masked}_{in}$ and the corresponding $\bm{m}$, we train the weight $\bm{W}$ of the network.
  In addition, $\bm{x}$ is not always available for all the modalities.
  For example, there may be a situation where $\bm{x}_{\{1, 2, 4\}}$ is available but $\bm{x}_{3}$ is not.
  In this case, for the mask of $\bm{x}$, $\bm{m}$ that can mask the data that cannot be obtained is chosen.
  If such $\bm{m}$ is not included in $\mathcal{M}$, we do not train using this data.
  For the loss function, the mean squared error is calculated only for the obtained data, and the weight $\bm{W}$ is updated.

  In addition, when parametric bias $\bm{p}$ is used as input, it is necessary to add one more step to the training method.
  In this case, we take data while changing the state of the body, tools, and environment.
  Let $D_{k}=\{\bm{x}_{1}, \bm{x}_{2}, \cdots, \bm{x}_{T_{k}}\}$ ($1 \leq k \leq K$) where $K$ is the total number of states and $T_{k}$ is the number of data in the state $k$.
  Thus, the data used for training is $D = \{(D_{1}, \bm{p}_{1}), (D_{2}, \bm{p}_{2}), \cdots, (D_{K}, \bm{p}_{K})\}$.
  Here, $\bm{p}_{k}$ is the parametric bias for the state $k$, which is a variable with a common value for the data $D_{k}$ but a different value for different data.
  Using this data $D$, we simultaneously update the network weight $\bm{W}$ and parametric bias $\bm{p}_{k}$.
  In other words, $\bm{p}_{k}$ is introduced so that the data $D_{k}$ of $\bm{x}$ in each state $k$ with different dynamics can be represented by a single network.
  This allows us to embed the dynamics of each state $k$ into $\bm{p}_{k}$, which can be applied to state recognition and adaptation to the current environment.
  Note that $\bm{p}_{k}$ is trained with an initial value of $\bm{0}$.
}%
{%
  本来はネットワーク構造の自動決定を先に行うが, その際にも本節のネットワーク学習が必要になるため, これを先に説明する.

  $\bm{x}$のデータ$D$が得られたとき, 通常であれば$\bm{x}_{in}$を入力として, 出力を推論し, 平均二乗誤差を損失関数としてこれが$\bm{x}_{out}$に近づくように学習を行う.
  一方, GeMuCoにはマスク変数$\bm{m}$が伴うため, これを学習に含める必要がある.
  まず, 実行可能な$\bm{m}$の集合$\mathcal{M}$を用意する.
  そして, 学習時には$\bm{x}_{in}$に対して$\mathcal{M}$に含まれるそれぞれの$\bm{m}$を使い, 対応する$\bm{x}_{in}$の一部をマスクし, $\bm{x}^{masked}_{in}$を作成する.
  この$\bm{x}^{masked}_{in}$と対応する$\bm{m}$を入力し, ネットワークの重み$\bm{W}$の学習を行う.
  また, $\bm{x}$は常に全てのモーダルのデータが得られているとは限らない.
  例えば, $\bm{x}_{\{1, 2, 4\}}$は得られるものの, 場合によって$\bm{x}_{3}$が得られない状態等が考えられる.
  このとき, この$\bm{x}$に対するマスクについては, 得られなかったデータをマスクできるような$\bm{m}$を選び用いる.
  そのような$\bm{m}$が$\mathcal{M}$に含まれなかった場合にはそのデータを使って学習は行わない.
  また損失関数についても, 得られたデータのみに対して平均二乗誤差を計算し, 重み$\bm{W}$を更新する.

  加えて, Parametric Bias $\bm{p}$を入力に用いる場合は, さらに一段学習方法を変化させる必要がある.
  この場合, 身体や道具・環境の状態を変化させながらデータを取る.
  ある状態$k$について得られたデータを$D_{k}=\{\bm{x}_{1}, \bm{x}_{2}, \cdots, \bm{x}_{T_{k}}\}$ ($1 \leq k \leq K$)とする($K$を全体の状態数, $T_{k}$を状態$k$におけるデータ数とする).
  よって, 学習に用いるデータは$D = \{(D_{1}, \bm{p}_{1}), (D_{2}, \bm{p}_{2}), \cdots, (D_{K}, \bm{p}_{K})\}$となる.
  ここで, $\bm{p}_{k}$はその試行$k$に関するParametric Biasであり, そのデータ$D_{k}$については共通の値で, 異なるデータについて別の値となる変数である.
  このデータ$D$を使って, ネットワークの重み$\bm{W}$とParametric Bias $\bm{p}_{k}$を同時に更新していく.
  つまり, ダイナミクスの異なるそれぞれの状態$k$における状態遷移データ$D_{k}$を一つのネットワークで表現できるように, $\bm{p}_{k}$を導入している.
  これにより, $\bm{p}_{k}$にそれぞれの状態$k$におけるダイナミクスが埋め込まれ, これを応用することで状態認識や環境適応が可能になる.
  なお, $\bm{p}_{k}$は初期値を0として学習される.
}%

\begin{figure}[t]
  \centering
  \includegraphics[width=1.0\columnwidth]{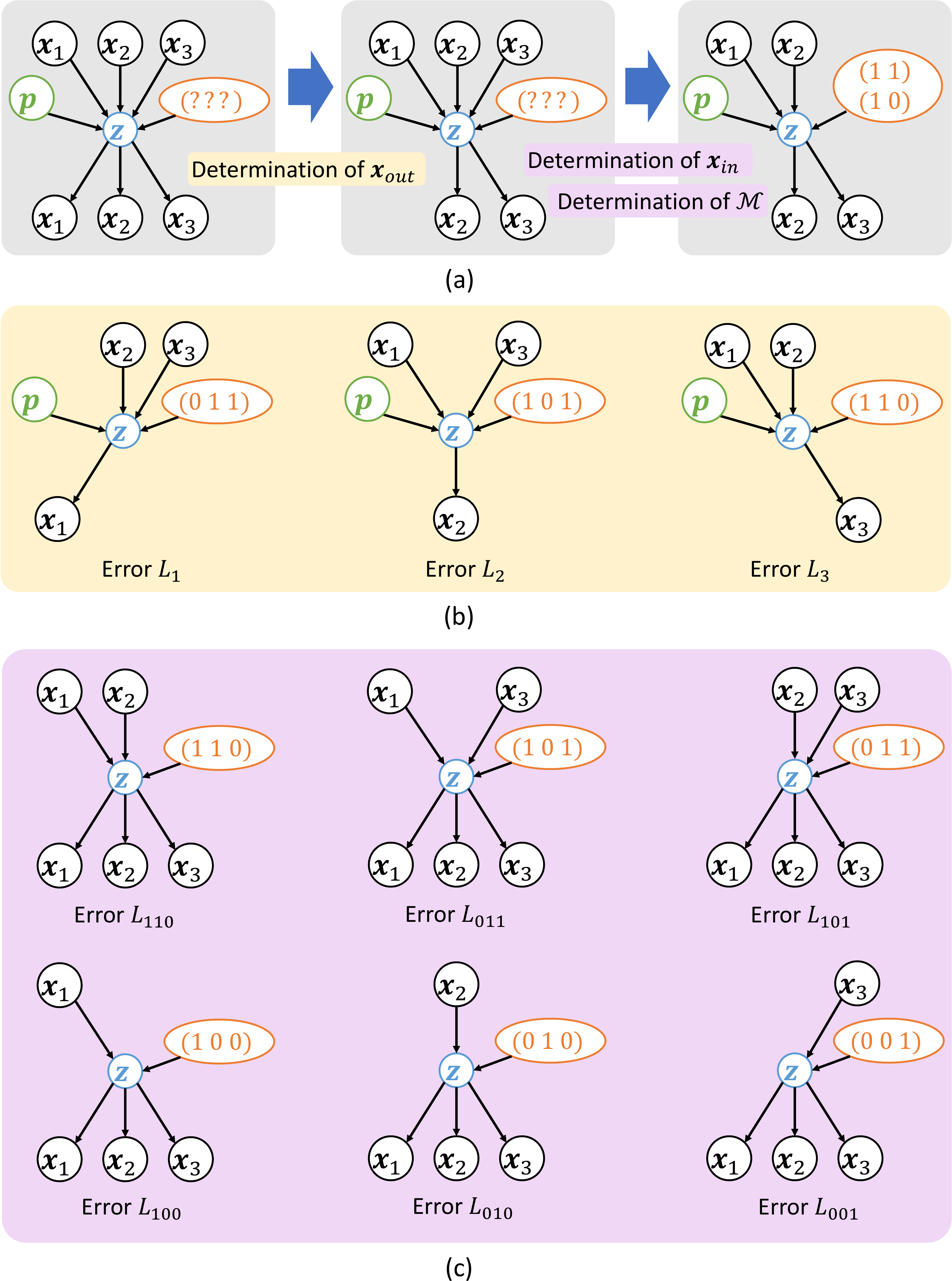}
  \caption{(a) The overview of the automatic determination of network input/output of GeMuCo. (b) The loss definition to determine the network output $\bm{x}_{out}$ of GeMuCo. (c) The loss definition to determine the network input $\bm{x}_{in}$ and a set of feasible masks $\mathcal{M}$ of GeMuCo.}
  \label{figure:inout-concept}
\end{figure}

\subsection{Automatic Structure Determination of GeMuCo} \label{subsec:structure-determinator}
\switchlanguage%
{%
  We describe a method for automatically determining the network structure of GeMuCo.
  Specifically, we determine $\bm{x}_{in}$, $\bm{x}_{out}$, and a set of feasible masks $\mathcal{M}$.
  If this network structure can be determined automatically from the given data, not only human work can be reduced, but also the autonomy of the robot can be dramatically improved.
  In other words, the robot can autonomously determine and train the network structure from the obtained data, and automatically construct state estimators, controllers, and so on based on the trained network structure.
  This operation mainly consists of determining network outputs that can be inferred from the latent space, and determining combinations of network inputs and masks that can infer the latent space, as shown in (a) of \figref{figure:inout-concept}.
  Note that the number of layers and units of the network are given externally by humans, and these are not automatically determined (there are various mechanisms such as NAS for these \cite{zoph2017nas}).

  \subsubsection{\textbf{Network Training}}
  In order to determine $\bm{x}_{in}$, $\bm{x}_{out}$ and a set of feasible masks $\mathcal{M}$, GeMuCo $\bm{h}$ is trained once using the obtained data $D$.
  The input/output of the network is determined based on the inference error when using the trained $\bm{h}$.
  Here, in order to calculate the inference error for each mask $\bm{m}$, when training the network as in \secref{subsec:network-trainer}, a set of all possible masks $\mathcal{M}_{all}$ (all $2^{N_{sensor}}-1$ combinations excluding masks that are all zero) is used.
  $\bm{m}$ is randomly selected from $\mathcal{M}_{all}$ each time.

  \subsubsection{\textbf{Determination of Network Output}}
  We determine the network output $\bm{x}_{out}$ of GeMuCo.
  This can be judged from whether a given value $\bm{x}_{i}$ is deducible from other values $\bm{x}_{j}$ ($i \neq j$).
  If it is deducible, then $\bm{x}_{i}$ is related to other sensors and actuators, and should be inferred as an output of the network.
  On the other hand, if it is not deducible, it should not be inferred because it will negatively influence the training of the network.
  As shown in (b) of \figref{figure:inout-concept}, a value $\bm{x}_{i}$ is inferred from other values $\bm{x}_{j}$ and its inference error is $L_{i}$.
  We collect only $\bm{x}_{i}$ for which $L_{i}<C^{out}_{thre}$, and construct $\bm{x}_{out}$ using them.
  Sensor values not adopted here are not utilized as part of the network output.

  \subsubsection{\textbf{Determination of Network Input}}
  We determine the network input $\bm{x}_{in}$ of GeMuCo.
  At the same time, we also determine a set of feasible masks $\mathcal{M}$ for the network input.
  This can be judged by the degree to which the value of $\bm{x}_{out}$ determined in the previous procedure can be inferred from each mask.
  The masks that allow inference, i.e. the combinations of $\bm{x}_{i}$ that allow inference, are extracted, and the set of such $\bm{x}_{i}$ becomes $\bm{x}_{in}$.
  If all inference errors are large when using the mask $\bm{m}$ containing a certain $\bm{x}_{i}$, then the $\bm{x}_{i}$ should be removed from $\bm{x}_{in}$. % TODO What's really going on? Shouldn't we take independent sets?
  First, we calculate the inference error $L_{m}$ of $\bm{x}_{out}$ for all $\bm{m}$.
  Let $\mathcal{X}_{out}$ be the set of sensors included in $\bm{x}_{out}$.
  Here, it is obvious that $\bm{x}_{out}$ can be inferred by the mask $\bm{m}$ corresponding to the set of sensors $\mathcal{X}$ such that $\mathcal{X}_{out}\subseteq\mathcal{X}$.
  Therefore, $L_{m}$ is calculated only for the set of sensors $\mathcal{X}$ such that $\mathcal{X}_{out}\nsubseteq\mathcal{X}$.
  For example, in (c) of \figref{figure:inout-concept}, $\mathcal{X}=\{\bm{x}_{1}, \bm{x}_{2}, \bm{x}_{3}\}$ is excluded from the calculation process since $\mathcal{X}_{out}=\{\bm{x}_{1}, \bm{x}_{2}, \bm{x}_{3}\}$.
  We collect $\bm{m}$ and the corresponding $\bm{x}_{i}$ for which $L_{m}<C^{in}_{thre}$, and denote their union set as $\mathcal{M}$ and $\bm{x}_{in}$, respectively.

  The automatic network input/output determination of GeMuCo depends on the threshold values of $C^{out}_{thre}$ and $C^{in}_{thre}$.
  Depending on the choice of these thresholds, the network structure changes, and the possible operations such as control and state estimation change accordingly.
  There is a tradeoff where setting low thresholds increase the accuracy of inference, but decrease the number of possible operations by reducing the number of input/output sensors.
  To mitigate this trade-off, there is a possibility of incorporating the accuracy of inference into the network by considering the output as both mean and variance.
  However, this aspect is not addressed in this study.
}%
{%
  GeMuCoのネットワーク構造を自動決定する方法について述べる.
  具体的には, $\bm{x}_{in}$, $\bm{x}_{out}$, そして実行可能なマスク$\bm{m}$の集合$\mathcal{M}$を決定する.
  このネットワーク構造を, 与えられたデータから自動で決定することができれば, 人間の作業を削減することができるだけでなく, ロボットの自律性を飛躍的に向上させることができる.
  つまり, 得られたデータから自律的にネットワーク構造を決定・学習し, これをもとに状態推定器や制御器等を自動構築するという一連の流れが可能になる.
  本操作は主に, 潜在空間から推論可能なネットワーク出力の決定, 潜在空間を推論可能なネットワーク入力とマスクの組み合わせの決定からなる(\figref{figure:inout-concept}の(a)).
  なお, ネットワークの層数やユニット数は人間が外から与えられる仕組みとし, これらについては自動決定しない(これらについては, NASなどの多様な仕組みが存在する\cite{zoph2017nas}).

  \subsubsection{ネットワーク学習}
  $\bm{x}$から$\bm{x}_{in}$, $\bm{x}_{out}$, 実行可能な$\bm{m}$の集合$\mathcal{M}$を決定するにあたり, 得られたデータ$D$をもとに一度GeMuCo $\bm{h}$を学習させる.
  この学習された$\bm{h}$を使った際の推論誤差を元に入出力が決定される.
  ここで, それぞれのマスク$\bm{m}$に対して推論誤差を計算するため, \secref{subsec:network-trainer}の学習の際には, 実行できる可能性のある全てのマスクの組み合わせ$\mathcal{M}_{all}$ (全て0であるマスクを除いた$2^{N_{sensor}}-1$の全組み合わせ)から毎回$\bm{m}$をランダムに取り出して用いる.

  \subsubsection{ネットワーク出力決定}
  GeMuCoのネットワーク出力$\bm{x}_{out}$を決定する.
  これは, ある値$\bm{x}_{i}$が, その他の値$\bm{x}_{j}$ ($i \neq j$)から推論可能であるかどうかから判断することができる.
  もし推論可能であればこの$\bm{x}_{i}$は他のセンサ・アクチュエータと関係しあっており, ネットワークの出力としても推論すべき対象である.
  一方, 推論可能でなければネットワークの学習に悪い影響を及ぼすため推論すべきではない.
  \figref{figure:inout-concept}の(b)に示すように, ある値$\bm{x}_{i}$をそれ以外の値$\bm{x}_{j}$から推論し, その誤差を$L_{i}$とする.
  $L_{i}<C^{out}_{thre}$である$\bm{x}_{i}$のみ集め, $\bm{x}_{out}$を構築する.
  ここで採用されなかったセンサ値はネットワーク出力としては用いない.

  \subsubsection{ネットワーク入力決定}
  GeMuCoのネットワーク入力$\bm{x}_{in}$を決定する.
  また, これと同時にネットワーク入力に対する実行可能なマスク$\bm{m}$の集合$\mathcal{M}$も決定する.
  これは, 前節で決定した$\bm{x}_{out}$の値を, それぞれのマスクがどの程度推論可能かどうかによって判断することができる.
  推論可能であったマスク, つまり推論を可能にする$\bm{x}_{i}$の組み合わせを抜き出し, それらの集合が$\bm{x}_{in}$となる.
  もしある$\bm{x}_{i}$を含むマスク$\bm{m}$を使った際の全ての推論誤差が大きい場合は, その$\bm{x}_{i}$は$\bm{x}_{in}$から削除すべきである. % TODO ほんとはどうなのよ. 独立な集合を取ったほうがいいんじゃないの？
  まず, 全ての$\bm{m}$に関して$\bm{x}_{out}$の推論誤差$L_{m}$を計算する.
  ここで, $\bm{x}_{out}$に含まれるセンサの集合を$\mathcal{X}_{out}$とすると, $\mathcal{X}_{out}\subseteq\mathcal{X}$となるセンサの集合$\mathcal{X}$と対応するマスク$\bm{m}$によって$\bm{x}_{out}$が推論できるのは自明である.
  ゆえに, $\mathcal{X}_{out}\nsubseteq\mathcal{X}$となるようなセンサの集合$\mathcal{X}$についてのみ, $L_{m}$は計算する.
  例えば, \figref{figure:inout-concept}の(c)においては, $\mathcal{X}_{out}=\{\bm{x}_{1}, \bm{x}_{2}, \bm{x}_{3}\}$であるため, $\mathcal{X}=\{\bm{x}_{1}, \bm{x}_{2}, \bm{x}_{3}\}$は除外されている.
  $L_{m}<C^{in}_{thre}$である$\bm{m}$とそれに対応する$\bm{x}_{i}$を集め, これらの和集合をそれぞれ$\mathcal{M}$と$\bm{x}_{in}$とする.

  これらネットワークの自動入出力決定は, $C^{out}_{thre}$と$C^{in}_{thre}$の閾値によって変化する.
  これらの選び方次第でネットワーク構造は変化し, それに応じて制御や状態推定等について可能な操作も変化する.
  これらの閾値を低く設定すると推論の精度は上がる一方, 入出力のセンサは減り, 可能な操作は減少するというトレードオフがある.
  なお, ネットワーク出力を平均と分散とすることで, 推論の精度自体をネットワークに取り込める可能性はあるが, 本研究では扱わない.
}%

\subsection{Online Update of GeMuCo} \label{subsec:online-updater}
\switchlanguage%
{%
  When the robot's physical state, tools, or surrounding environment changes, a model adapted to the current state can be used by updating GeMuCo for accurate state estimation and control.
  There are three possible ways to update the network: updating $\bm{W}$, updating $\bm{p}$, and updating $\bm{W}$ and $\bm{p}$ simultaneously.
  This corresponds to (c) and (d), or a combination of them in \figref{figure:basic-usage}.
  When data $D$ is obtained, the loss function is computed and the gradient descent method is used to update only $\bm{W}$, only $\bm{p}$, or $\bm{W}$ and $\bm{p}$ simultaneously.
  In the case of offline update, the network is updated once after a certain amount of data has been accumulated.
  In the case of online update, the network is updated gradually.
  When the number of data exceeds a determined threshold, data is discarded from the oldest.
  In addition to the actual $D$, if there are any constraints such as origin, geometric model, minimum and maximum values, etc., the data representing these constraints can also be added to $D$ during the training.
  In the case of updating only $\bm{p}$, only some dynamics are changed and the structure of the overall network is kept the same, thus overfitting for the current data is unlikely to occur.
  On the other hand, it should be noted that updating $\bm{W}$ or updating $\bm{W}$ and $\bm{p}$ simultaneously changes the structure of the entire network, and thus overfitting is likely to occur.
}%
{%
  ロボットの身体状態や道具, 周囲環境が変化した際, ネットワークを更新しその状態に適応したモデルを用いることで, 正確な状態推定や制御が可能になる.
  ネットワークの更新には$\bm{W}$の更新, $\bm{p}$の更新, $\bm{W}$と$\bm{p}$の同時更新の3種類が考えられる.
  これは, \figref{figure:basic-usage}における(c)と(d), またはその組み合わせに相当する.
  データ$D$が得られたとき, 損失関数を計算し, この誤差を使って勾配法により$\bm{W}$のみ, $\bm{p}$のみ, または$\bm{W}$と$\bm{p}$を同時に更新する.
  オフライン更新の際には, ある一定のデータが溜まった後に一度にこの更新を行う.
  一方, オンライン更新の際には, ある閾値よりもデータが溜まってから逐次的にネットワークを更新していく.
  データの最大数を決め, それよりも多くデータが集まった場合は古いものから順にデータを捨てていく.
  学習の際, 実際に得られた$D$に加えて, 原点や幾何モデル, 最小値や最大値等何らかの制約がある場合には, これを表現するデータも$D$に加えて更新を行う.
  $\bm{p}$のみの更新では, ネットワーク全体の構造は変化せず一部のダイナミクスのみが変化するため, 現在のデータに対する過学習が起きにくい.
  一方, $\bm{W}$の更新や$\bm{W}$と$\bm{p}$の同時更新ではネットワーク全体の構造が変化するため, 過学習が起きやすい点には注意が必要である.
}%

\subsection{Optimization Computation by Iterative Forward and Backward Propagations} \label{subsec:optimization-computation}
\switchlanguage%
{%
  We describe the optimization computation that is frequently used in the state estimation, control, and simulation, which will be explained subsequently.
  This operation corresponds to (e) and (f) of \figref{figure:basic-usage}, where the procedure of iterative optimization of $\bm{x}$ or $\bm{z}$ based on a certain loss function is shown.
  As an example, let us assume that $\bm{z}$ is optimized based on the loss function for $\bm{x}_{out}$ and that there is the relation $\bm{x}_{out} = \bm{h}_{dec}(\bm{z})$.
  \begin{enumerate}
    \renewcommand{\labelenumi}{\arabic{enumi})}
    \item Assign the initial value $\bm{z}^{init}$ to the variable $\bm{z}^{opt}$ to be optimized
    \item Infer the predicted value of $\bm{x}_{out}$ as $\bm{x}^{pred}_{out}= \bm{h}_{dec}(\bm{z}^{opt})$
    \item Calculate the loss $L$ using the loss function $\bm{h}_{loss}$
    \item Calculate $\partial{L}/\partial{\bm{z}^{opt}}$ using the backward propagation
    \item Update $\bm{z}^{opt}$ by gradient descent method
    \item Repeat processes 2)--5) to optimize $\bm{z}^{opt}$
  \end{enumerate}
  We now describe process 5) in detail.
  Process 5) performs the following operation,
  \begin{align}
    \bm{z}^{opt} \gets \bm{z}^{opt} - \gamma\frac{\partial{L}}{\partial{\bm{z}^{opt}}} \label{eq:optimization-computation}
  \end{align}
  where $\gamma$ is the learning rate.
  $\gamma$ can be a constant, but we can also try various $\gamma$ as variables to achieve faster convergence.
  For example, we determine the maximum value $\gamma_{max}$ of $\gamma$, divide $[0, \gamma_{max}]$ equally into $N_{batch}$ values ($N_{batch}$ is a constant that expresses the batch size of training), and update $\bm{z}^{opt}$ with each $\gamma$.
  Then, we select $\bm{z}^{opt}$ with the smallest $L$ in steps 2) and 3), and repeat steps 4) and 5) with various $\gamma$ for the $\bm{z}^{opt}$.
}%
{%
  後の状態推定・制御・シミュレーションで多用する最適化計算について述べる.
  これは\figref{figure:basic-usage}の(e), (f)に相当する操作であり, ある損失関数をもとに$\bm{x}$や$\bm{z}$を繰り返し最適化する手順を以下に示す.
  ここでは例として, $\bm{z}$を$\bm{x}_{out}$に関する損失関数をもとに最適化することとし, $\bm{x}_{out} = \bm{h}_{dec}(\bm{z})$の関係があるとする.
  \begin{enumerate}
    \renewcommand{\labelenumi}{(\arabic{enumi})}
    \item 最適化される変数$\bm{z}^{opt}$に初期値$\bm{z}^{init}$を代入する.
    \item $\bm{x}_{out}$の予測値$\bm{x}^{pred}_{out}= \bm{h}_{dec}(\bm{z}^{opt})$を推論する.
    \item 損失関数$\bm{h}_{loss}$を使い, 損失$L$を計算する.
    \item 誤差逆伝播により, $\partial{L}/\partial{\bm{z}^{opt}}$を計算する.
    \item 勾配法により, $\bm{z}^{opt}$を更新する.
    \item (2)--(5)の工程を繰り返し, $\bm{z}^{opt}$を最適化する.
  \end{enumerate}
  ここで, (5)の工程について詳しく説明する.
  (5)の工程は基本的に以下のような操作を行う.
  \begin{align}
    \bm{z}^{opt} \gets \bm{z}^{opt} - \gamma\frac{\partial{L}}{\partial{\bm{z}^{opt}}} \label{eq:optimization-computation}
  \end{align}
  ここで, $\gamma$は学習率である.
  この$\gamma$は定数でも良いが, 変数として様々な$\gamma$を試し, より速い収束を目指すこともできる.
  例えば, $\gamma$の最大値$\gamma_{max}$を決め, $[0, \gamma_{max}]$を$N_{batch}$等分し($N_{batch}$は学習のバッチサイズとなる定数), それぞれの$\gamma$で$\bm{z}^{opt}$を更新する.
  その後, (2)と(3)の工程で最も$L$の小さかった$\bm{z}^{opt}$を選び, これに対して(4)と(5)の工程を行うことを繰り返す.
}%

\subsection{State Estimation using GeMuCo} \label{subsec:state-estimator}
\switchlanguage%
{%
  In state estimation, the sensor values that are currently unavailable are estimated from the network.
  For this purpose, (a), (b), (e), and (f) in \figref{figure:basic-usage} are used.
  If $\bm{x}_{out}$ contains the value to be estimated, we consider the execution of (a) or (b), and if (a) and (b) are not possible, we consider the execution of (e).
  If $\bm{x}_{out}$ does not contain the value to be estimated, we consider the execution of (f).

  In (a), we consider a mask $\bm{m}$, which is set to $0$ for the unavailable data and $1$ for the available data.
  If this mask $\bm{m}$ is included in the set of feasible masks $\mathcal{M}$, then by inputting this $\bm{m}$ and $\bm{x}^{masked}_{in}$ with $\bm{0}$ for the unavailable data into the network, we can estimate the currently unavailable data.

  Similarly, in (b), if the network has all necessary inputs, the remaining data can be estimated directly.
  Even when the inputs are not available, if there exists a feasible $\bm{m}$, the missing data can be estimated in the same way as in (a).

  If there is no feasible $\bm{m}$ in the form of (a) and (b), state estimation is performed in the form of (e).
  This corresponds to the case where the loss function is set as follows in \secref{subsec:optimization-computation},
  \begin{align}
    \bm{h}_{loss}(\bm{x}^{pred}_{out}, \bm{x}^{data}_{out}) = ||\bm{m}_{x_{out}}\odot(\bm{x}^{pred}_{out}-\bm{x}^{data}_{out})||_{2} \label{eq:state-loss}
  \end{align}
  where $\bm{x}^{data}_{out}$ is the currently obtained data of $\bm{x}_{out}$ with some unavailable data.
  Also, $\bm{m}_{x_{out}}$ is a mask that is $0$ for unavailable data and $1$ for available data.
  $\odot$ denotes Hadamard product, and $||\cdot||_{2}$ denotes L2 norm.
  In other words, this procedure corresponds to updating $\bm{z}^{opt}$ so that the predictions are consistent only for the obtained data.
  The obtained $\bm{x}^{pred}_{out}$ is used as the estimated state $\bm{x}^{est}_{out}$.

  If $\bm{x}_{out}$ does not contain the value to be estimated, state estimation is performed in the form (f).
  This corresponds to the case in \secref{subsec:optimization-computation} where the variable to be optimized $\bm{z}^{opt}$ and its initial value $\bm{z}^{init}$ are changed to $\bm{x}^{opt}_{in}$ and $\bm{x}^{init}_{in}$, respectively.
  The loss function is the same as \equref{eq:state-loss}.
  That is, instead of the latent representation $\bm{z}$, we propagate the error directly to the network input $\bm{x}_{in}$ and use the obtained $\bm{x}^{opt}_{in}$ as the estimated state $\bm{x}^{est}_{in}$.
}%
{%
  状態推定は, 現在得られないセンサ値をネットワークから推定することを意味する.
  これには\figref{figure:basic-usage}における(a), (b), (e), (f)が用いられる.
  $\bm{x}_{out}$に推定したい値が含まれる場合, (a)または(b)の実行を考え, これらが実行出来ない場合に(e)の実行を考える.
  $\bm{x}_{out}$に推定したい値が含まれない場合, (f)の実行を考える.

  (a)において, 現在得られていないデータについて0, 得られているデータについて1としたマスク$\bm{m}$を考える.
  もしこのマスク$\bm{m}$が実行可能なマスクの集合$\mathcal{M}$に含まれていれば, 得られていないデータを$\bm{0}$とした$\bm{x}^{masked}_{in}$とこの$\bm{m}$をネットワークに入力することで, 得られていないデータを推定することが可能である.

  (b)においても同様で, ネットワークに必要な入力が揃っている場合には, 直接残りのデータを推定することができる.
  入力が揃っていない場合でも, 実行可能な$\bm{m}$が存在すれば, (a)と同様の形で得られていないデータを推定することができる.

  (a)と(b)の形で実行可能な$\bm{m}$が存在しない場合は, (e)の形で状態推定を行う.
  これは, \secref{subsec:optimization-computation}において, 損失関数を以下のように設定した場合に相当する.
  \begin{align}
    \bm{h}_{loss}(\bm{x}^{pred}_{out}, \bm{x}^{data}_{out}) = ||\bm{m}_{x_{out}}\odot(\bm{x}^{pred}_{out}-\bm{x}^{data}_{out})||_{2}
  \end{align}
  ここで, $\bm{x}^{data}_{out}$は現在得られた$\bm{x}_{out}$のデータであり, 得られなかった一部のデータが欠損している.
  また, $\bm{m}_{x_{out}}$は欠損しているデータについて0, していないデータについて1を取るマスクである.
  $\odot$はアダマール積, $||\cdot||_{2}$はL2ノルムを表している.
  つまり, 得られたデータについてのみ予測が合致するように$\bm{z}^{opt}$を更新していくことに相当する.
  最終的に, 得られた$\bm{z}^{opt}$から$\bm{x}^{opt}_{out}$を推定する.

  $\bm{x}_{out}$に推定したい値が含まれない場合は(f)の形で状態推定を行う.
  これは, \secref{subsec:optimization-computation}において, 最適化される変数$\bm{z}^{opt}$とその初期値$\bm{z}^{init}$が, それぞれ$\bm{x}^{opt}_{in}$と$\bm{x}^{init}_{in}$に変更された場合に相当する.
  損失関数は前述の(e)の場合と同様である.
  つまり, 潜在表現$\bm{z}$ではなく, 直接ネットワーク入力$\bm{x}_{in}$に誤差を伝播し, 得られた$\bm{x}^{opt}_{in}$を推定値として用いる.
}%

\subsection{Control Using GeMuCo} \label{subsec:control}
\switchlanguage%
{%
  Depending on the structure of the network, either (a), (b), (e), or (f) in \figref{figure:basic-usage} is computed, as in \secref{subsec:state-estimator}.
  The calculation depends on whether the control input is contained in $\bm{x}_{in}$ or $\bm{x}_{out}$, and whether the target state can be input directly or the target state needs to be expressed in the form of a loss function.

  If the control input is contained in $\bm{x}_{out}$ and all target states can be input directly from $\bm{x}_{in}$, then either (a) or (b) is performed.
  That is, $\bm{x}_{out} = \bm{h}(\bm{x}_{in}, \bm{m})$.

  If the control input is contained in $\bm{x}_{out}$ and the target state must be expressed in the form of a loss function, (e) is executed.
  This corresponds to the case where the loss function is executed in the form of $\bm{h}_{loss}(\bm{x}^{pred}_{out}, \bm{x}^{ref}_{out})$ in \secref{subsec:optimization-computation}.
  Here, $\bm{x}^{ref}_{out}$ is the target state of $\bm{x}_{out}$, and $\bm{h}_{loss}$ can take various forms, e.g. $||A\bm{x}^{pred}_{1}-\bm{x}^{ref}_{1}||_{2}$, $||\bm{x}^{pred}_{1}-\bm{x}^{ref}_{1}||_{2}+||\bm{x}^{pred}_{2}||_{2}$, etc. ($\bm{x}^{\{pred, ref\}}_{\{1, 2\}}$ represents the $\{1, 2\}$-th sensor value in $\bm{x}^{\{pred, ref\}}_{out}$, and $A$ denotes a certain transformation matrix).
  $\bm{z}^{opt}$ is optimized from this loss function, and the obtained $\bm{x}^{pred}_{out}$ is used as the control input.

  If the control input is not included in $\bm{x}_{out}$, (f) is executed.
  This means that the variable $\bm{z}^{opt}$ to be optimized and its initial value $\bm{z}^{init}$ in \secref{subsec:optimization-computation} are changed to $\bm{x}^{opt}_{in}$ and $\bm{x}^{init}_{in}$, respectively.
  Since the loss function can also include the loss with respect to $\bm{x}^{opt}_{in}$, the loss function is $\bm{h}_{loss}(\bm{x}^{pred}_{out}, \bm{x}^{ref}_{out}, \bm{x}^{opt}_{in})$, unlike \equref{eq:state-loss}.
  In other words, the error is propagated directly to the network input $\bm{x}_{in}$ instead of the latent representation $\bm{z}$.
  The obtained $\bm{x}^{opt}_{in}$ is used as the control input.
}%
{%
  制御はネットワークの構造次第で, \secref{subsec:state-estimator}と同様に, \figref{figure:basic-usage}における(a), (b), (e), (f)のいずれかの計算が行われる.
  制御入力が$\bm{x}_{in}$または$\bm{x}_{out}$のどちらに含まれるか, 指令状態を直接入力できる, または指令状態を損失関数の形で表現する必要があるかによってどの計算を用いるかが変わる.

  制御入力が$\bm{x}_{out}$に含まれ, 指令状態が全て$\bm{x}_{in}$から直接入力できる場合は, (a)または(b)が実行される.
  つまり, $\bm{x}_{out} = \bm{h}(\bm{x}_{in}, \bm{m})$となる.

  次に, 制御入力が$\bm{x}_{out}$に含まれ, 指令状態を損失関数の形で表現しなければいけない場合, (e)が実行される.
  これは, \secref{subsec:optimization-computation}において, 損失関数を$\bm{h}_{loss}(\bm{x}^{pred}_{out}, \bm{x}^{ref}_{out})$の形で実行した場合に相当する.
  $\bm{x}^{ref}_{out}$は$\bm{x}_{out}$の指令状態であり, $\bm{h}_{loss}$は例として, $||A\bm{x}^{pred}_{1}-\bm{x}^{ref}_{1}||_{2}$や$||\bm{x}^{pred}_{1}-\bm{x}^{ref}_{1}||_{2}+||\bm{x}^{pred}_{2}||_{2}$等, 様々な形が考えられる($\bm{x}^{\{pred, ref\}}_{\{1, 2\}}$は$\bm{x}^{\{pred, ref\}}_{out}$における$\{1, 2\}$番目のセンサ値を表し, $A$は何らかの変換行列を表す).
  これを元に$\bm{z}^{opt}$を計算し, 最終的に得られた$\bm{x}^{opt}_{out}$から制御入力を計算, 実機に指令する.

  最後に, 制御入力が$\bm{x}_{out}$に含まれない場合には(f)が実行される.
  これは, \secref{subsec:optimization-computation}において, 最適化される変数$\bm{z}^{opt}$とその初期値$\bm{z}^{init}$が, それぞれ$\bm{x}^{opt}_{in}$と$\bm{x}^{init}_{in}$に変更された場合に相当する.
  損失関数は$\bm{x}^{opt}_{in}$に関する損失も含むことができるため, 前述の(e)と異なり$\bm{h}_{loss}(\bm{x}^{pred}_{out}, \bm{x}^{ref}_{out}, \bm{x}^{opt}_{in})$となる.
  つまり, 潜在表現$\bm{z}$ではなく, 直接ネットワーク入力$\bm{x}_{in}$に誤差を伝播する.
  得られた$\bm{x}^{opt}_{in}$から制御入力を計算, 実機に指令する.
}%

\subsection{Simulation using GeMuCo} \label{subsec:simulation}
\switchlanguage%
{%
  In simulation, the current robot state is estimated from the control input and some constraints.
  The simulation can be performed in the form of (a), (b), (e), or (f), which is almost the same as the state estimation in \secref{subsec:state-estimator}.
  The only difference is the loss functions in (e) and (f).
  Since the loss function does not have the current data $\bm{x}^{data}_{out}$ as in the state estimation but instead has the control input value $\bm{x}^{send}_{out}$ that is actually commanded, the loss function becomes $\bm{h}_{loss}(\bm{x}^{pred}_{out}, \bm{x}^{send}_{out})$.
  This loss function describes various constraints on the motion of the robot, such as joint torque, muscle tension, and motion speed.
  Based on the control input and the constraints given in the form of a loss function, the current state is estimated and transitioned.
}%
{%
  シミュレーションとは, 制御入力といくつかの拘束条件から現在のロボット状態を推定することを意味する.
  \secref{subsec:state-estimator}における状態推定とほぼ同じ形で, (a), (b), (e), (f)のいずれかの形でシミュレーションを行うことができる.
  このとき異なるのは(e), (f)における損失関数のみである.
  損失関数は状態推定のように現在のデータ$\bm{x}^{data}_{out}$が存在しない代わりに実際に送っている指令値$\bm{x}^{send}_{out}$が得られるため, $\bm{h}_{loss}(\bm{x}^{pred}_{out}, \bm{x}^{send}_{out})$の形で記述される.
  この損失関数はロボットの動作に関する関節トルクや筋張力, 動作速度等に関する様々な拘束を記述することになる.
  制御入力と損失関数の形で与えられた拘束をもとに, 現在状態を遷移させていく.
}%

\subsection{Anomaly Detection using GeMuCo} \label{subsec:anomaly-detection}
\switchlanguage%
{%
  Anomaly detection is performed with respect to the amount of error between the current value $\bm{x}_{out}$ and the estimated value $\bm{x}^{est}_{out}$.
  One of the simplest anomaly detection methods is to set a threshold value for $||\bm{x}_{out}-\bm{x}^{est}_{out}||_{2}$ and consider an anomaly when the error is larger than the threshold.
  On the other hand, the mean and variance of the error can be used to detect anomalies more accurately.
  First, we collect the state estimation data $\bm{x}^{est}_{out}$ and the current state data $\bm{x}^{data}_{out}$ in the normal state without any anomaly.
  For this data, we calculate the mean $\bm{\mu}$ and variance $\bm{\Sigma}$ of the error $\bm{e}^{data}_{out}=\bm{x}^{data}_{out}-\bm{x}^{est}_{out}$.
  When actually detecting an anomaly, the difference $\bm{e}_{out}$ between the current value $\bm{x}_{out}$ and the estimated value $\bm{x}^{est}_{out}$ is always obtained, and the Mahalanobis distance $d = \sqrt{(\bm{e}_{out}-\bm{\mu})^{T}\bm{\Sigma}^{-1}(\bm{e}_{out}-\bm{\mu})}$ is calculated for it.
  When $d$ exceeds the threshold value, we assume that an anomaly has been detected.
  % For this threshold value, it is possible to calculate the variance of $d$ for the data obtained when calculating $\bm{\mu}$ and $\bm{\Sigma}$, and use the 3-sigma interval or the like.
}%
{%
  異常検知は$\bm{x}_{out}$に関する現在値$\bm{x}_{out}$と推定値$\bm{x}^{est}_{out}$の誤差の大きさに対して行う.
  一つの最も単純な異常検知方法は, $||\bm{x}_{out}-\bm{x}^{est}_{out}||_{2}$に対して閾値を設定し, これよりも誤差が大きくなった場合に異常と見なす方法である.
  一方, その誤差の平均と分散を使うことで, より正確に異常検知を行うことができる.
  まず, 異常のない, 正常な状態において\secref{subsec:state-estimator}の状態推定のデータ$\bm{x}^{est}_{out}$と現在状態データ$\bm{x}^{data}_{out}$を収集する.
  このデータに対して, 誤差$\bm{e}^{data}_{out}=\bm{x}^{data}_{out}-\bm{x}^{est}_{out}$の平均$\bm{\mu}$と分散$\bm{\Sigma}$を計算しておく.
  実際に異常検知をする際には, $\bm{x}_{out}$に関する現在値$\bm{x}_{out}$と推定値$\bm{x}^{est}_{out}$の差$\bm{e}_{out}$を常に取得し, これに対して以下のマハラノビス距離$d$を計算する.
  \begin{align}
    d = \sqrt{(\bm{e}_{out}-\bm{\mu})^{T}\bm{\Sigma}^{-1}(\bm{e}_{out}-\bm{\mu})}
  \end{align}
  この$d$が設定した閾値を超えたとき, 異常が検知されたこととする.
  この閾値であるが, $\bm{\mu}$と$\bm{\Sigma}$を計算する際に得られたデータに対して$d$の分散を計算しておき, その3シグマ区間等を用いることが可能である.
}%

\begin{figure*}[t]
  \centering
  \includegraphics[width=1.95\columnwidth]{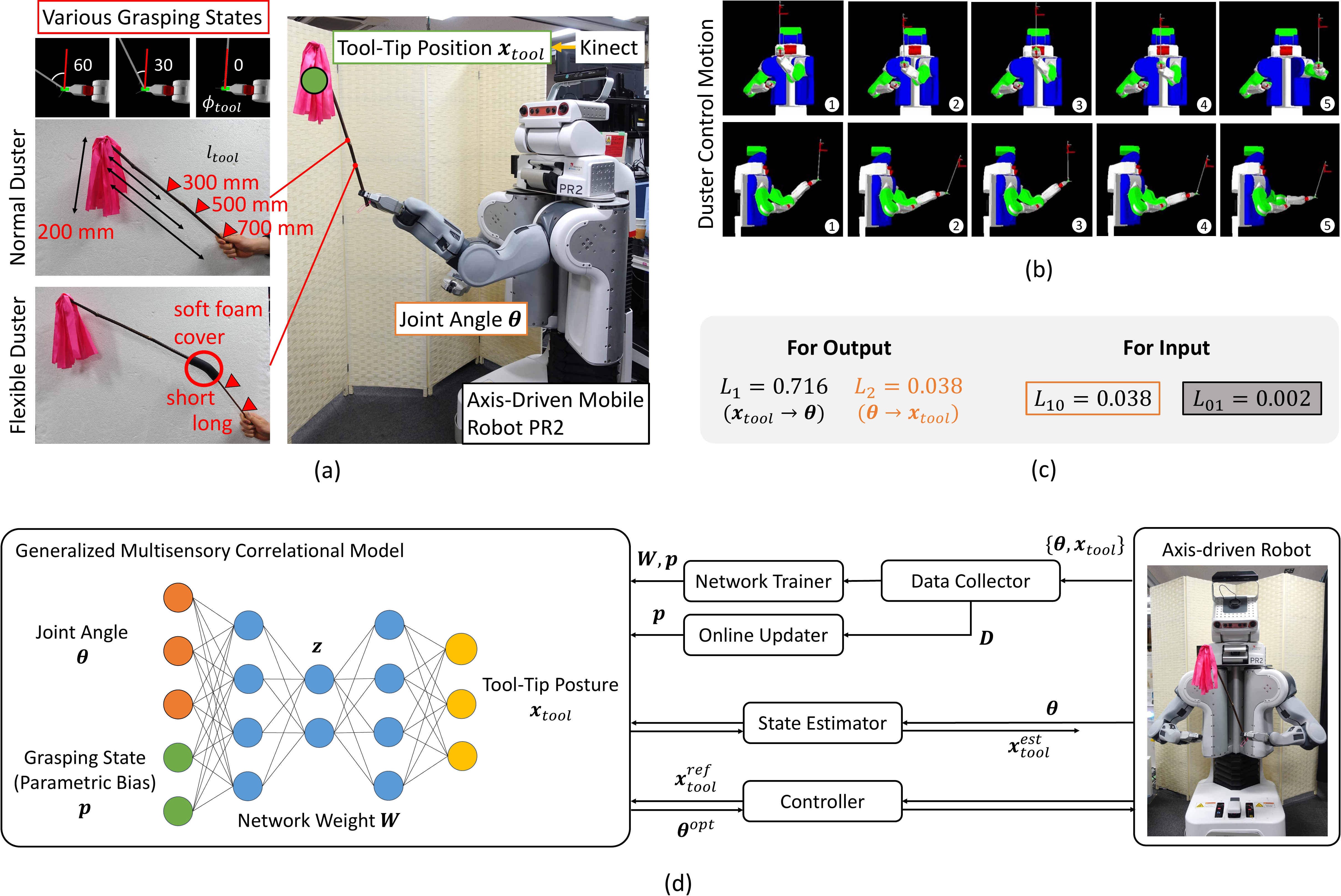}
  \caption{(a) Adaptive tool-tip control learning considering online changes in grasping state. We prepared two types of dusters, Normal Duster and Flexible Duster, with various grasping states (grasping angle and position). The basic operation of dusters are shown in the bottom half. (b) The automatic determination of network structure for adaptive tool-tip control learning. (c) The system configuration for adaptive tool-tip control learning.}
  \label{figure:pr2-overview}
\end{figure*}

\section{Experiments} \label{sec:experiment}
\switchlanguage%
{%
  In this study, we utilize the proposed body schema model of GeMuCo for (i) adaptive tool-tip control learning considering the change in grasping state, (ii) complex tendon-driven body control learning for musculoskeletal humanoids, and (iii) full-body tool manipulation learning for low-rigidity humanoids.
  (i) uses the simplest network structure, and due to its simplicity, no mask expression is required after the network structure is automatically determined.
  (ii) uses a more complex network structure but without parametric bias, in which the entire network is trained online.
  (iii) uses the most complex network structure, which contains all the elements of GeMuCo.
  These are newly interpreted experiments of \cite{kawaharazuka2021tooluse, kawaharazuka2020autoencoder, kawaharazuka2024kxr} using the GeMuCo framework.

  The reason for adopting each network structure is as below.
  When utilizing parametric bias, it is necessary to collect data while varying the state.
  Therefore, parametric bias is employed in cases where it is easy to vary the state, such as in (i) and (iii) scenarios involving changes in grasping state or tool conditions.
  On the other hand, for aspects such as correction of sim-to-real gap and body changes due to aging or deterioration, as in (ii), it is challenging to capture data while varying the state.
  Therefore, parametric bias is not used in such cases.
  Additionally, depending on the task, the collection of sensor values and the automatic determination of network input and output result in the determination of the network structure.

  For the automatic determination of the network structure, $C^{\{out, in\}}_{thre}$ is generally set to $\{0.15, 0.15\}$.
  In (ii), we also analyze the case where $C^{\{out, in\}}_{thre}=\{0.3, 0.3\}$ in the sense of increasing the number of possible network operations.
}%
{%
  本研究では, 提案するGeMuCoによる身体図式を用いて, (i)把持状態変化を考慮した適応的道具先端操作学習, (ii)筋骨格ヒューマノイドの身体制御学習, (iii)低剛性ヒューマノイドの全身道具操作制御学習を行う.
  (i)は最も単純なネットワーク構造であり, シンプルゆえにネットワーク構造の自動決定後はマスク表現が不要な形である.
  (ii)はより複雑なネットワーク構造であるが, Parametric Biasは除きネットワーク全体をオンライン学習した例である.
  (iii)はGeMuCoの全ての要素が詰まった最も複雑な制御を行っている.
  (i)と(ii)は\cite{kawaharazuka2021tooluse}と\cite{kawaharazuka2020autoencoder}をGeMuCoのフレームワークとして新しく捉え直した形である.

  なお, 各ネットワーク構造を採用した理由ですが, Parametric Biasを使う場合は, 状態を変化させながらデータを取る必要があるため, (i)や(iii)のように把持状態や道具の変化など, 状態を簡単に変化させてデータを取得しやすい場合にParametric Biasを採用しています.
  一方で(ii)のような, シミュレーションと実機の誤差の修正や, 経年劣化に基づく身体変化については, 状態を変化させながらデータを取得することが難しいため, Parametric Biasは用いていません.
  また, そのタスクに応じて, 得られるセンサ値を収集し, ネットワークの自動入出力決定を行った結果として, そのネットワーク構造が決まっています.

  なお, ネットワーク構造の自動決定について, $C^{\{out, in\}}_{thre}$は基本的に$\{0.15, 0.15\}$と設定しているが, (ii)についてはネットワークが可能な操作を増やす意味で$\{0.3, 0.3\}$とした場合について解析している.
}%

\subsection{Adaptive Tool-Tip Control Learning Considering Online Changes in Grasping State}
\switchlanguage%
{%
  Various studies have been conducted on tool manipulation, but they do not take into consideration the fact that the grasping position and angle of a tool changes gradually during tool manipulation.
  Moreover, few studies have dealt with deformable tools, and generally, only rigid tools fixed to the body have been considered.
  In this experiment, we propose a body schema learning method to control the tip position of rigid and flexible tools by considering gradual changes in the grasping state.
}%
{%
  道具操作については様々な研究が行われてきたが, それらは道具操作中に逐次的に道具の把持位置・角度等が変化してしまうことを考慮していない.
  また, 変形する道具を扱っている研究も少なく, 基本的には剛体の道具が身体に固定された状態のみを扱っている.
  そこで本実験では, 提案する身体図式学習により, ロボットが逐次的な把持状態変化を考慮しながら剛な道具や柔軟な道具の先端位置を制御することを行う.
}%

\begin{figure*}[t]
  \centering
  \includegraphics[width=2.0\columnwidth]{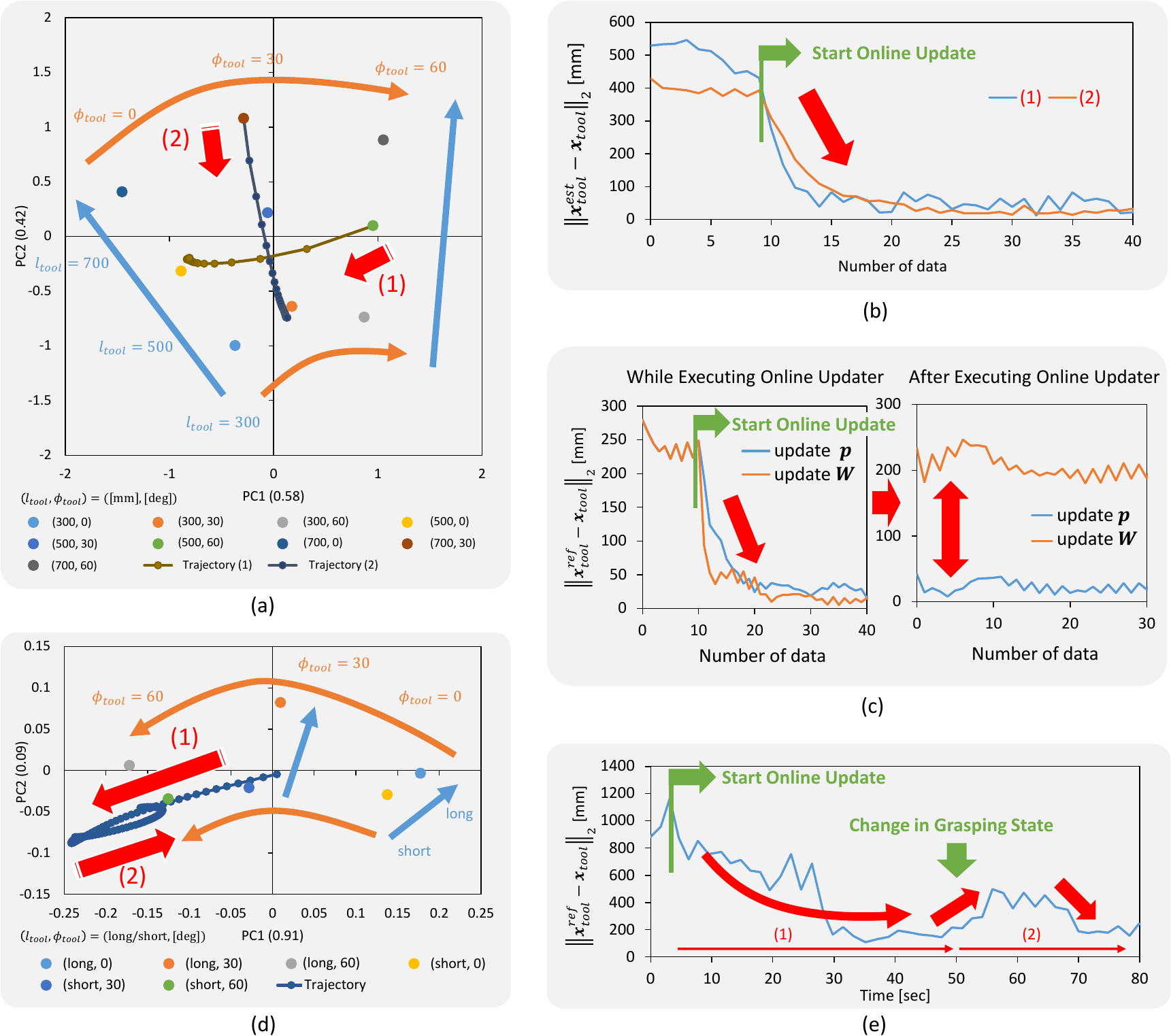}
  \caption{(a) The trained parametric bias and its trajectory during online update of GeMuCo in the simulation experiment. (b) The transition of the state estimation error during online update of GeMuCo in the simulation experiment. (c) The transition of the control error while and after executing two types of online updaters, ``update $\bm{p}$'' and ``update $\bm{W}$'', in the simulation experiment. (d) The trained parametric bias and its trajectory during online update of GeMuCo in the actual robot experiment. (e) The transition of the control error during online update of GeMuCo in the actual robot experiment \cite{kawaharazuka2021tooluse}.}
  \label{figure:pr2-exp}
\end{figure*}

\subsubsection{\textbf{Experimental Setup}}
\switchlanguage%
{%
  We conduct experiments using a duster as shown in (a) of \figref{figure:pr2-overview}.
  A duster is a tool to remove dust from shelves and crevices by controlling the tool-tip position.
  In this experiment, we set $\bm{x}=\{\bm{x}_{tool}, \bm{\theta}\}$.
  Here, $\bm{x}_{tool}$ is the tool-tip position and $\bm{\theta}$ is the 7-dimensional joint angle used as the control input to the robot.

  Experiments using a simulation and the actual machine of the mobile robot PR2 are conducted.
  In the simulation experiment, we prepare an artificial cylinder (Normal Duster) to represent a duster.
  Since the cloth of the duster hangs down from the tip of the stick in the direction of gravity, the tool-tip position $\bm{x}_{tool}$ is assumed to be 100 mm below the tip of the stick in the simulation.
  The data is obtained by changing the grasping position of the duster (expressed as the length of the tool from the hand) $l_{tool}$ and the grasping angle (angle perpendicular to the parallel gripper in one degree of freedom) $\phi_{tool}$ in three different ways: $l_{tool}=\{300, 500, 700\}$ [mm] and $\phi_{tool}=\{0, 30, 60\}$ [deg].

  In the actual robot experiment, we handle a more difficult situation using a Flexible Duster, in which a Normal Duster and an additional stick are connected by a flexible foam material, and the tool-tip position changes significantly depending on the angle at which the duster is held.
  The length of the duster used in the experiment is 500 mm, the length of the cloth is 200 mm, and the length of the additional stick is 250 mm.
  % The head of PR2 is equipped with a depth camera, Kinect (Miscrosoft, Corp.).
  We perform Euclidean clustering of the color-extracted points of the cloth of the duster, and the center position of the largest cluster is the tool-tip position.
  As in the simulation experiment, $l_{tool}$ and $\phi_{tool}$ are varied in the actual robot.
  Since the grasping state is implicitly trained even if the parameters of the grasping state are not directly known, the data is collected by manually and roughly creating grasping states with long/short tool length and $\phi_{tool}=\{0, 30, 60\}$.

  Note that $\bm{p}$ is set to be two-dimensional in each experiment.
}%
{%
  本研究では, 掃除用具の一つであるはたきを使った実験を行う(\figref{figure:pr2-overview}の(a)).
  はたきはその道具の先端位置を制御することで, 棚や隙間のほこりを落とす道具である.
  本研究では$\bm{x}=\{\bm{x}_{tool}, \bm{\theta}\}$としている.
  ここで, $\bm{x}_{tool}$は道具の先端位置, $\bm{\theta}$はロボットへの制御入力で7次元の関節角度である.

  本研究の実験では, 台車型ロボットPR2のシミュレーションと実機を用いて実験を行う.
  シミュレーション実験では, 擬似的に真っ直ぐな棒(Normal Duster)をはたきとして用意する.
  はたきの布は棒の先端から重力方向に垂れるため, 道具先端位置$\bm{x}_{tool}$は棒の先端から重力方向に100 mm下の場所としてシミュレーションを行う.
  はたきの把持位置(手先からの道具の長さで表現) $l_{tool}$, 把持角度(平行グリッパと垂直方向に関する角度を一自由度で表現) $\phi_{tool}$を3種類ずつ変化させながらデータを取得する.
  なお, $l_{tool}=\{300, 500, 700\}$ [mm], $\phi_{tool}=\{0, 30, 60\}$ [deg]とした.

  実機実験では, より難しい問題として, はたきと追加の棒を発泡性の柔軟素材により接続し, 持つ角度によって道具先端位置が大きく変化するはたき(Flexible Duster)を用いた実験を行う.
  なお, 実機で用いるはたきは棒の長さが500 mm, 布は200 mm, 追加の棒は250 mmである.
  PR2の頭部には深度カメラであるKinect (Miscrosoft, Corp.)がついており, 色抽出した道具の先端の点群をEuclidean Clusteringし, 一番大きなClusterの中心位置を道具の先端位置とした.
  シミュレーションと同様に実機でも$l_{tool}$と$\phi_{tool}$を変化させる.
  なお, 把持状態は直接把持状態のパラメータが分からなくても暗黙的に学習されるため, 人間が目分量で, 道具を長く持った状態(long)と短く持った状態(short), $\phi_{tool}=\{0, 30, 60\}$の状態を作り, データを収集している.

  なお, それぞれの実験で$\bm{p}$は2次元とした.
}%

\subsubsection{\textbf{Network Structure}}
\switchlanguage%
{%
  In the simulation, the joint angle is moved randomly, and 1000 data points per grasping state are collected, which amounts to 9000 data points in total.
  The automatic determination of the network structure based on the obtained data is shown in (c) of \figref{figure:pr2-overview}.
  First, we compute $L_{\{1, 2\}}$ for determining the network output.
  While $L_{2}$ is small, $L_{1}$ is larger than $C^{out}_{thre}$ and is not deducible.
  Therefore, the only output of the network is $\bm{x}_{tool}$.
  Next, we compute $L_{m}$ for determining the network input.
  Since the output is only $\bm{x}_{tool}$, $L_{\{01, 11\}}$ is not computed.
  Therefore, only $L_{10}$ is computed, and as it is small enough ($L_{10}$ is smaller than $C^{in}_{thre}$), the input is set to $\bm{\theta}$.
  The network $\bm{\theta}\rightarrow\bm{x}_{tool}$ is constructed and the system configuration shown in (d) of \figref{figure:pr2-overview} is automatically constructed.
}%
{%
  シミュレーションにおいて関節角度をランダムに動かし, 一つの把持状態に対して1000, 全体で9000のデータを取得した.
  得られたデータに基づくネットワーク構造の自動決定について\figref{figure:pr2-overview}の(c)に示す.
  まず, ネットワーク出力の決定のための$L_{\{1, 2\}}$を計算した.
  $L_{2}$は小さいのに対して, $L_{1}$は$C^{out}_{thre}$より大きく, 推論できるとは言いがたい.
  よって, ネットワークの出力は$\bm{x}_{tool}$のみとなった.
  次に, ネットワーク入力決定のための$L_{m}$を計算した.
  出力が$\bm{x}_{tool}$のみであるため, $L_{\{01, 11\}}$は計算されない.
  よって, $L_{10}$のみが計算され, これが十分小さいと判断し, 入力は$\bm{\theta}$となった.
  よって, $\bm{\theta}\rightarrow\bm{x}_{tool}$というネットワークが構築され, \figref{figure:pr2-overview}の(d)に示すようなシステム構成が自動構築される.
}%

\subsubsection{\textbf{Simulation Experiment}}
\switchlanguage%
{%
  % We perform a duster operation experiment in simulation.
  First, we randomly move the joint angle in the simulation to obtain 1000 data points per grasping state, amounting to 9000 data points in total, and train the network based on these data points.
  The trained parametric bias $p_k$ is represented in two-dimensional space through Principle Component Analysis (PCA) as shown in (a) of \figref{figure:pr2-exp}.
  We can see that each PB is neatly self-organized along the size of $l_{tool}$ and $\phi_{tool}$.
  The larger $l_{tool}$ is, the larger the difference of PB with the change of $\phi_{tool}$ is, which is consistent with the point that the tool-tip position changes more significantly with the grasping angle for longer tools.

  Next, we test the behavior of online update of parametric bias and tool-tip position estimation.
  Experiments are performed for two cases in which the grasping state is changed from (1) $(l_{tool}, \phi_{tool})=(500, 60)$ to $(l_{tool}, \phi_{tool})=(500, 0)$ or (2) $(l_{tool}, \phi_{tool})=(700, 30)$ to $(l_{tool}, \phi_{tool})=(300, 30)$.
  The duster control motion is shown in (b) of \figref{figure:pr2-overview} (with a tool at $(l_{tool}, \phi_{tool})=(500, 30)$).
  The target tool-tip position $\bm{x}^{ref}_{tool}$ is repeatedly moved (200, -200) [mm] in the $(x, z)$ direction and then moved back while advancing by 100 mm in the $y$ direction from a certain reference point.
  The trajectories of parametric bias, ``Trajectory (1)'' and ``Trajectory (2)'', during this duster motion are shown in (a) of \figref{figure:pr2-exp}, and the error of the estimated tool-tip position $||\bm{x}^{est}_{tool}-\bm{x}_{tool}||_{2}$ is shown in (b) of \figref{figure:pr2-exp} ($\bm{x}^{est}_{tool}$ refers to the estimated value of the tool-tip position).
  For both (1) and (2), we can see that the parametric bias is gradually approaching the current grasping state obtained at training phase.
  We can also see that the estimation error of the tool-tip position gradually decreases accordingly.
  The averaged estimation errors for (1) and (2) are 52.2 mm and 25.9 mm, respectively, when more than 20 data points were collected.

  Finally, we experiment with the control of the tool-tip position.
  When the parametric bias is initially set to $(l_{tool}, \phi_{tool})=(500, 30)$ obtained at training phase and then set to $(l_{tool}, \phi_{tool})=(500, 60)$, we compare the control error when only $\bm{p}$ is updated online (update $\bm{p}$) to when $\bm{p}$ is fixed but $\bm{W}$ is updated (update $\bm{W}$).
  The former case updates only $\bm{p}$, while the latter corresponds to updating the weight $\bm{W}$ without $\bm{p}$, as in the usual online learning.
  The duster control motion is the same as in (b) of \figref{figure:pr2-overview}.
  Here, in order to use the joint angle $\bm{\theta}^{orig}$ of (b) of \figref{figure:pr2-overview} generated as $(l_{tool}, \phi_{tool})=(500, 30)$, we set the loss as follows,
  \begin{align}
    L = ||\bm{x}^{pred}_{tool}-\bm{x}^{ref}_{tool}||_{2} + 0.3||\bm{\theta}^{opt}-\bm{\theta}^{orig}||_{2}
  \end{align}
  where $\bm{x}^{pred}_{tool}$ is the predicted value of $\bm{x}_{tool}$ and $\bm{\theta}^{opt}$ is $\bm{\theta}$ to be optimized.
  The transition of the control error of the tool-tip position $||\bm{x}^{ref}_{tool}-\bm{x}_{tool}||_{2}$ is shown in the left figure of (c) of \figref{figure:pr2-exp}.
  It can be seen that the control error is about 240 mm in the initial period when the online updater does not work, while the control error is significantly reduced by the online updater.
  When the number of data points obtained is more than 20, the average control error is 31.5 mm for ``update $\bm{p}$'' and 19.2 mm for ``update $\bm{W}$''.
  The latter, which updates the entire network, is more accurate.
  The right figure of (c) of \figref{figure:pr2-exp} shows the transition of the control error when the online updater is stopped and the same tool-tip position is realized by a different $\bm{\theta}^{orig}$ with a different tool-tip rotation constraint.
  After updating only $\bm{p}$, the control error is 22.6 mm on average, while it is 207 mm after updating $\bm{W}$.
  It is found that the online update of grasping state is effective for other joint angles when only $\bm{p}$ is updated, while the control error increases for other joint angles when $\bm{W}$ is updated due to overfitting on the data used for training.
}%
{%
  シミュレーション上において, はたき操作実験を行う.
  先と同様にシミュレーションにおいて関節角度をランダムに動かし, 一つの把持状態に対して1000, 全体で9000のデータを取得し, これを元にネットワークを学習させる.
  このときに得られたParametric Bias $p_k$をPCAを通して2次元空間に表現した図を\figref{figure:pr2-exp}の(a)に示す.
  $l_{tool}$と$\phi_{tool}$の大小に伴って, それぞれのPBが規則的に整列していることがわかる.
  $l_{tool}$が大きいほど$\phi_{tool}$の変化によるPBの違いが大きくなっており, より長い道具ほど角度によって大きく道具先端位置が変化する点と一致している.

  次に, Parametric Biasのオンライン学習と道具先端位置推定の挙動について実験する.
  道具を(1) $(l_{tool}, \phi_{tool})=(500, 60)$から$(l_{tool}, \phi_{tool})=(500, 0)$に変化させた場合, (2) $(l_{tool}, \phi_{tool})=(700, 30)$から$(l_{tool}, \phi_{tool})=(300, 30)$に変化させた場合の2つの場合について実験を行う.
  はたきを振る動作は, \figref{figure:pr2-overview}の(b)に示すようなものである($(l_{tool}, \phi_{tool})=(500, 30)$の道具を持った場合).
  道具先端の指令位置$\bm{x}^{ref}_{tool}$は, 基準点を決め, そこからy方向に100 mmずつ進みながら, (x, z)方向に(200, -200) [mm]動かしては戻すという動作を繰り返す.
  この動きの際のParametric Biasの動き(``Trajectory (1)''と``Trajectory (2)'')を\figref{figure:pr2-exp}の(a)に, 道具先端位置の推定誤差$||\bm{x}^{est}_{tool}-\bm{x}_{tool}||_{2}$の遷移を\figref{figure:pr2-exp}の(b)に示す($\bm{x}^{est}_{tool}$は道具先端位置の推定値を指す).
  (1), (2)の両者について, Parametric Biasは訓練時に得られた現在の把持状態周辺に徐々に近づいていることがわかる.
  また, それに伴い, 道具先端位置の推定誤差も徐々に下がっていくことがわかる.
  データが20以上集まった時における推定誤差の平均は(1)が52.2 mm, (2)は25.9 mmであった.

  最後に, 道具先端位置の制御について実験する.
  Parametric Biasが訓練時に得られた$(l_{tool}, \phi_{tool})=(500, 30)$である状態から始め, $(l_{tool}, \phi_{tool})=(500, 60)$としたときに, $\bm{p}$のみをオンライン学習した場合(update $\bm{p}$)と, $\bm{p}$は固定して$\bm{W}$を更新した場合(update $\bm{W}$)での制御誤差を比較する.
  前者は$\bm{p}$のみを更新するのに対して, 後者は通常のオンライン学習のように, $\bm{p}$は存在せずに重み$\bm{W}$を更新していくことに相当する.
  動作は\figref{figure:pr2-overview}の(b)と同じ動作である.
  このとき, $(l_{tool}, \phi_{tool})=(500, 30)$として生成された\figref{figure:pr2-overview}の(b)の関節角度$\bm{\theta}^{orig}$を基準とするため, 損失を以下のように設定する.
  \begin{align}
    L = ||\bm{x}^{pred}_{tool}-\bm{x}^{ref}_{tool}||_{2} + 0.3||\bm{\theta}^{opt}-\bm{\theta}^{orig}||_{2}
  \end{align}
  ここで, $\bm{x}^{pred}_{tool}$の$\bm{x}_{tool}$のネットワーク予測値, $\bm{\theta}^{opt}$は$\bm{\theta}$の最適化変数を表す.
  道具先端位置の制御誤差$||\bm{x}^{ref}_{tool}-\bm{x}_{tool}||_{2}$の遷移を\figref{figure:pr2-exp}の(c)の左図に示す.
  online updaterが働かない初期は制御誤差が約240 mmなのに対して, online updaterによって制御誤差が大きく下がっていることがわかる.
  得られたデータ数が20以上となったとき, 制御誤差の平均は, update $\bm{p}$で31.5 mm, update $\bm{W}$で19.2 mmであり, ネットワーク全体を更新する後者の方が精度が高かった.
  その後online updaterを止め, 道具先端の回転制約を変えて先と異なる$\bm{\theta}^{orig}$で同じ道具先端位置を制御した際の制御誤差の遷移を\figref{figure:pr2-exp}の(c)の右図に示す.
  $\bm{p}$のみ更新した後では, 制御誤差は平均で22.6 mmなのに対して, $\bm{W}$を更新した後では207 mmであった.
  $\bm{p}$のみを更新した場合は他の関節角度においても把持状態更新が効果を発揮するのに対して, $\bm{W}$を更新した場合は訓練に使ったデータに過学習してしまい, 他の関節角度においては制御誤差が大きくなってしまうことがわかった.
}%

\subsubsection{\textbf{Actual Robot Experiment}}
\switchlanguage%
{%
  The actual robot experiment is performed using PR2 with Flexible Duster.
  We perform the same operation with the simulation as shown in (b) of \figref{figure:pr2-overview} three times while changing the reference point.
  We repeat the above motions while changing the grasping state, and train GeMuCo using the approximately 1500 data points that were obtained.
  At this time, the model obtained from the simulation described above is fine-tuned.
  The trained parametric bias is shown in (d) of \figref{figure:pr2-exp}.
  Since the Flexible Duster is more bendable than the Normal Duster, the parametric bias varies much more depending on the grasping angle than the grasping position (tool length).

  The results of the tool-tip control experiment similar to the simulation experiment in (c) of \figref{figure:pr2-exp} are shown in (e) of \figref{figure:pr2-exp}.
  The initial control error is very large (about 1000 mm), but the error is gradually reduced to about 150 mm as the current grasping state is recognized.
  When a human applies external force to the tool to change the grasping state, the control error increases to about 500 mm, but the error is reduced again to about 200 mm by online updater.
  The transitions of PB are shown in ``Trajectory'' of (d) of \figref{figure:pr2-exp}, where (1) is the transition just after the start of online updater and (2) is the transition after the change of the grasping state.
  It can be seen that PB is autonomously updated by detecting the change of the grasping state.
}%
{%
  PR2を用いた実機実験を行う.
  前述のシミュレーションで行った\figref{figure:pr2-overview}の(b)の動作を, 基準点を変えながら3回行いデータを取得する.
  把持状態を変化させながら上記動作を繰り返し, 得られた約1500のデータを使いGeMuCoを学習させる.
  このとき, 前述のシミュレーションで得られたモデルをFine-Tuningする.
  学習によって得られたParametric Biasを\figref{figure:pr2-exp}の(d)に示す.
  Flexible DusterはNormal Dusterよりもさらに持つ角度によって道具自体が大きく曲がるため, Parametric Biasがshort/longに比べて$\phi_{tool}$によって非常に大きく変化する形となっている.

  \figref{figure:pr2-exp}の(c)と同様の道具先端制御実験を行った結果を\figref{figure:pr2-exp}の(e)に示す.
  初期の制御誤差は約1000 mmと非常に大きいが, 徐々に把持状態が分かっていき, 150 mm程度まで誤差が減っている.
  その後人間が道具に外力を加えて把持状態を変更すると500 mm程度まで制御誤差は増えるが, Online Updaterによりまた200 mm程度まで下がった.
  このときのPBの遷移は\figref{figure:pr2-exp}の(d)の``Trajectory''の通りであり, (1)がOnline State Updater開始後, (2)が把持状態の変化後の遷移である.
  把持状態の変化を検知して自動でPBが更新されていることがわかる.
}%

\subsection{Complex Tendon-driven Body Control Learning for Musculoskeletal Humanoids}
\switchlanguage%
{%
  We perform body schema learning that can handle state estimation, control, and simulation of musculoskeletal humanoids in a unified manner.
  By updating this network online based on the sensor data of the actual robot, the state estimation, control, and simulation of musculoskeletal humanoids can be performed more accurately and continuously.
  Note that the control is based on muscle length-based control, and dynamic factors including hysteresis due to high friction between muscles and bones are not handled in this experiment, resulting in handling only static relationships.
  In addition, since parametric bias is not used in this study, we do not obtain data for various body states.
}%
{%
  筋骨格ヒューマノイドの状態推定・制御・シミュレーションを統一的に扱うことができる身体図式学習を行う.
  本ネットワークを実機センサデータをもとにオンラインで更新していくことで, 筋骨格ヒューマノイドの状態推定・制御・シミュレーションをより正確かつ継続的に行うことができるようになる.
  なお, 制御は筋長制御ベースであり, 本実験ではヒステリシスを含む動的な要素については扱わない.
  また, 本研究ではParametric Biasは用いていないため, 多様な身体状態におけるデータは取得しない.
}%

\begin{figure*}[t]
  \centering
  \includegraphics[width=1.8\columnwidth]{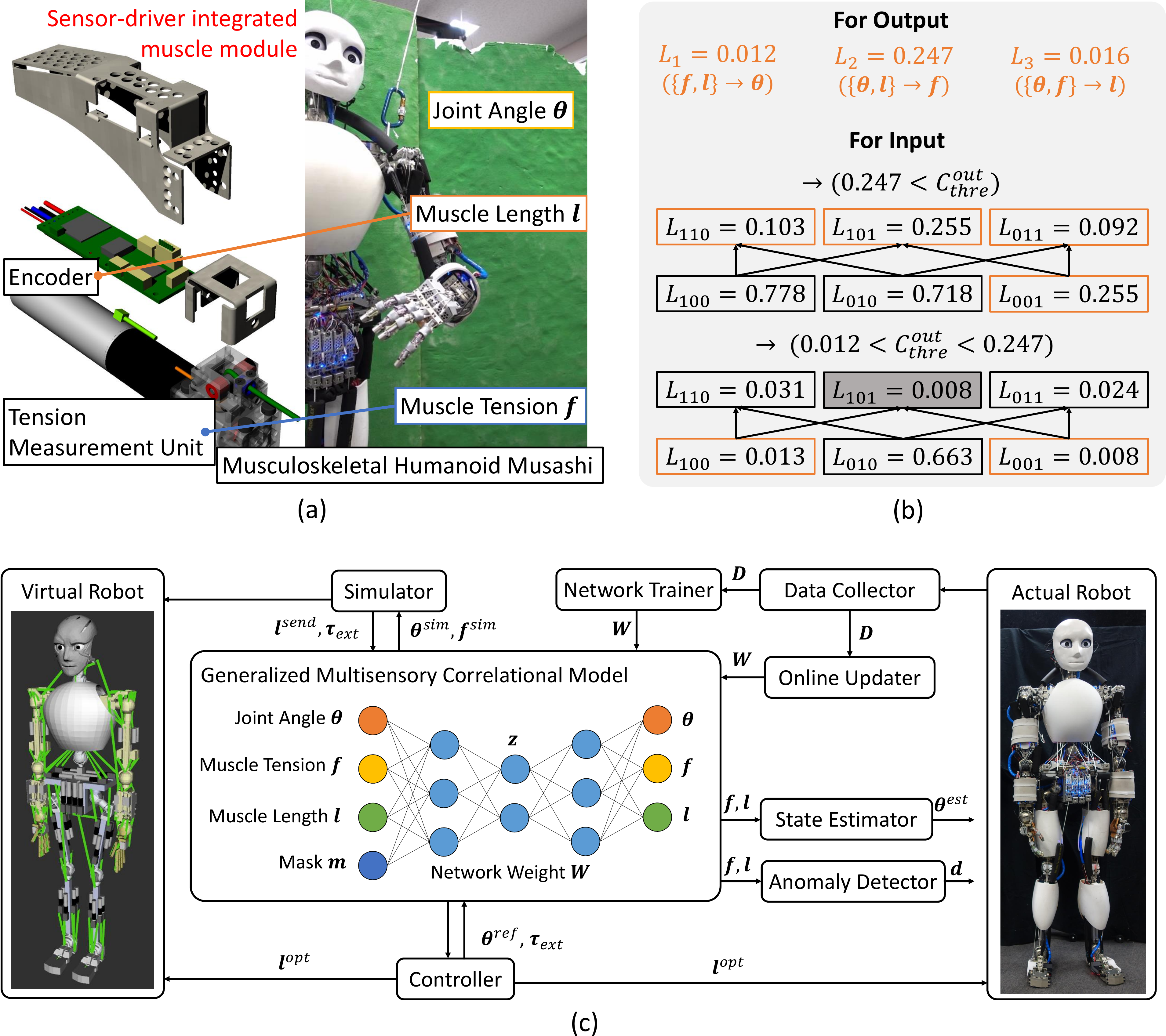}
  \caption{(a) Complex tendon-driven body control learning for musculoskeletal humanoids with sensors of joint angle, muscle tension, and muscle length. (b) Automatic determination of network structure for complex tendon-driven body control learning. (c) The system configuration for complex tendon-driven body control learning.}
  \label{figure:musashi-overview}
\end{figure*}

\subsubsection{\textbf{Experimental Setup}}
\switchlanguage%
{%
  In this experiment, we use a musculoskeletal humanoid Musashi \cite{kawaharazuka2019musashi} as shown in (a) of \figref{figure:musashi-overview}.
  We mainly use the 3 degrees of freedom (DOFs) of the shoulder and the 2 DOFs of the elbow as the joint angle $\bm{\theta}$. %, and the joint angle is expressed as $\bm{\theta}=(\theta_{S-r}, \theta_{S-p}, \theta_{S-y}, \theta_{E-p}, \theta_{E-y})$ ($S$ for the shoulder, $E$ for the elbow, and $rpy$ for the roll, pitch, and yaw, respectively).
  Ten muscles move the 5 DOFs of the shoulder and elbow, of which one muscle is a biarticular muscle.
  For each muscle, muscle length $l$ is obtained from an encoder and muscle tension $f$ from a loadcell.
  Although joint angle $\bm{\theta}$ of musculoskeletal humanoids cannot be directly obtained because their joints are usually ball joints, Musashi can measure the joint angle with its pseudo-ball joint module \cite{kawaharazuka2019musashi}.
  Even if the joint angle cannot be directly measured, $\bm{\theta}$ can be obtained by first estimating the rough joint angle based on muscle length changes and then correcting them using AR markers attached to the hand \cite{kawaharazuka2018online}.
  In this study, we set $\bm{x}=\{\bm{\theta}, \bm{f}, \bm{l}\}$.

  This musculoskeletal humanoid has a geometric model that is a representation of the muscle route by connecting the start, relay, and end points of the muscle with straight lines.
  Given a certain joint angle, muscle length can be obtained from the distance between the muscle relay points.
  By considering the elongation of the nonlinear elastic element attached to the muscle depending on muscle tension, the muscle length can be obtained from the given joint angle and the muscle tension.
  On the other hand, since it is difficult to simulate the wrapping of the muscle around the joint and its change over time, this geometric model is quite different from that of the actual robot and some learning using the actual robot sensor data is necessary.
}%
{%
  本実験では筋骨格ヒューマノイドMusashi \cite{kawaharazuka2019musashi}を用いる(\figref{figure:musashi-overview}の(a)).
  ここでは主に, 肩の3自由度, 肘の2自由度を用いて実験を行うこととし, 関節角度は$\bm{\theta}=(\theta_{S-r}, \theta_{S-p}, \theta_{S-y}, \theta_{E-p}, \theta_{E-y})$のように表す($S$はshoulder, $E$はelbow, $rpy$はそれぞれroll, pitch, yawを表す).
  筋はこの肩と肘の5自由度に対して10本配置されており, 2関節筋は1本である.
  それぞれの筋について, エンコーダから筋長$l$, ロードセルから筋張力$f$が得られる.
  通常筋骨格ヒューマノイドの関節は球関節のため関節角度$\bm{\theta}$を直接得ることは出来ないが, Musashiは擬似球関節モジュールにより関節角度を測定することが可能である.
  一方, 関節角度が得られない場合も手先にARマーカ等を取り付ければ筋長変化に基づく関節角度推定とマーカを用いた補正により関節角度$\bm{\theta}$を得ることができる\cite{kawaharazuka2018online}.
  本研究では$\bm{x}=\{\bm{\theta}, \bm{f}, \bm{l}\}$としている.

  この筋骨格ヒューマノイドは幾何モデルを持つ.
  ここで言う幾何モデルとは, 筋の起始点・中継点・終止点を直線で結び筋経路を表現したものである.
  関節角度を指定すると, 筋の経由点間の距離から筋長を求めることができる.
  ここに筋に付属する非線形弾性要素単体の張力に対する伸びを考慮することで, 指定した関節角度と筋張力から筋長を求めることができる.
  一方, 筋の関節に対する巻きつきや経年変化等を模擬することが難しいため, この幾何モデルは実機とは大きく異なり, 実機センサデータを用いた学習が必要である.
}%

\begin{figure*}[t]
  \centering
  \includegraphics[width=2.0\columnwidth]{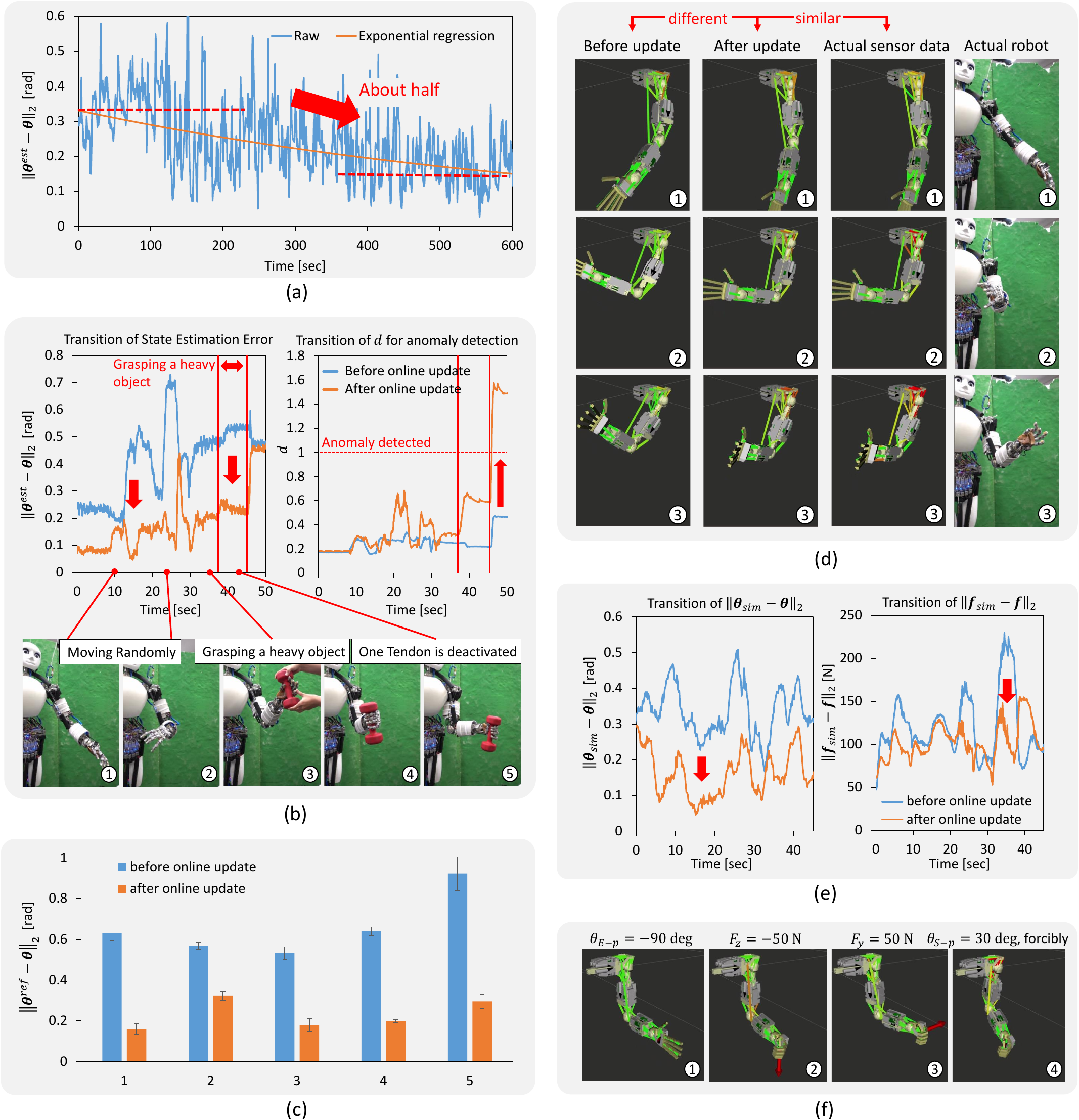}
  \caption{(a) The transition of state estimation error while executing online updater. (b) The transition of state estimation error and anomaly score before and after online update of GeMuCo. (c) The control error before and after online update of GeMuCo. (d) The difference between the actual robot motion and the simulated robot motion before and after online update of GeMuCo. (e) The transition of the simulated joint angle and muscle tension errors before and after online update of GeMuCo. (f) Modification of the simulation behavior by changing the loss function \cite{kawaharazuka2020autoencoder}.}
  \label{figure:musashi-exp}
\end{figure*}

\subsubsection{\textbf{Network Structure}}
\switchlanguage%
{%
  We obtain 10000 data points from the random joint angle and muscle tension movements of the simulation.
  The automatic determination of the network structure based on the obtained data is shown in (b) of \figref{figure:musashi-overview}.
  First, $L_{\{1, 2, 3\}}$ is computed to determine the network output.
  While $L_{1}$ and $L_{3}$ are almost equally small, $L_{2}$ is somewhat larger.
  We now consider the cases where $0.247<C^{out}_{thre}=0.30$, i.e. all sensors are used for output, and where $0.012<C^{out}_{thre}=0.15<0.247$, i.e. $\bm{f}$ is not used.
  For these cases, we compute $L_{m}$ for network input determination.
  In the former case, when $C^{in}_{thre}$ is set to 0.15, only $L_{110}$ and $L_{011}$ are less than this value, so $(\bm{\theta}, \bm{f}, \bm{l})$, the union of sensors used in these masks, is used as the input (note that since the output is $(\bm{\theta}, \bm{f}, \bm{l})$, $L_{111}$ is not calculated).
  This is a network structure whose input and output are $(\bm{\theta}, \bm{f}, \bm{l})$ as well.
  When $C^{in}_{thre}$ is set to 0.30, the number of mask types further increases from 2 to 4.
  In the latter case where $f$ is not used for the output, when $C^{in}_{thre}$ is set to about 0.015, only $L_{100}$ and $L_{001}$ are below this value, so $(\bm{\theta}, \bm{l})$, the union of the sensors used in these masks, is used as the input (note that since the output is $(\bm{\theta}, \bm{l})$, $L_{\{101, 111\}}$ is not calculated).
  This is a type of network that does not take muscle tension into account.
  In other words, networks for various musculoskeletal structures can be constructed only with these different threshold values.
  In this study, we use the former network structure in which muscle tension is taken into account to enable various operations of GeMuCo.
  The constructed system is shown in (c) of \figref{figure:musashi-overview}.
}%
{%
  シミュレーションのランダムな関節と筋張力の動きから10000のデータを取得する.
  得られたデータに基づくネットワーク構造の自動決定について\figref{figure:musashi-overview}の(b)に示す.
  まず, ネットワーク出力の決定のための$L_{\{1, 2, 3\}}$を計算した.
  $L_{1}$と$L_{3}$はほぼ同程度に小さいのに対して, $L_{2}$はそれらと比べると多少大きいが, 全体として損失自体は小さいと言える.
  ここで, $0.247<C^{out}_{thre}=0.30$, つまり全センサを出力に用いる場合と, $0.012<C^{out}_{thre}=0.15<0.247$, つまり$\bm{f}$が使われない場合について考える.
  これらに対して, ネットワーク入力決定のための$L_{m}$を計算した.
  前者では, 例えば$C^{in}_{thre}$を0.15に設定した場合, $L_{110}$と$L_{011}$のみがこれを下回るため, これらマスクに用いられたセンサの和集合である$(\bm{\theta}, \bm{f}, \bm{l})$全てが入力として使われる(なお, 出力が$(\bm{\theta}, \bm{f}, \bm{l})$であるため, $L_{111}$は計算されない).
  これは入出力が同様に$(\bm{\theta}, \bm{f}, \bm{l})$であるネットワーク構造である.
  また, $C^{in}_{thre}$を0.30に設定した場合, さらにマスクの種類が2から4に増える.
  後者では, 例えば$C^{in}_{thre}$を0.015程度に設定した場合, $L_{100}$と$L_{001}$のみがこれを下回るため, これらマスクに用いられたセンサの和集合である$(\bm{\theta}, \bm{l})$が入力として使われる(なお, 出力が$(\bm{\theta}, \bm{l})$であるため, $L_{\{101, 111\}}$は計算されない).
  これは, 筋張力を考慮しない形のネットワークである.
  つまり, これら閾値の違いのみで, 様々な筋骨格構造に関するネットワークが構築可能である.
  本研究では, 筋張力を考慮する前者のネットワーク構造を用いる.
  構築された全体システムを\figref{figure:musashi-overview}の(c)に示す.
}%

\subsubsection{\textbf{Actual Robot Experiment}}
\switchlanguage%
{%
  First, we perform initial training of the network using a geometric model.
  The network is initialized from 10000 data points by obtaining muscle length from random joint angle and muscle tension movements.
  Next, we perform online update of the model based on the actual robot sensor data.
  Here, the online update is performed by repeatedly moving the robot with random joint angles and muscle tensions.
  The transition of the error between the estimated joint angle $\bm{\theta}^{est}$ and the currently measured joint angle $\bm{\theta}$, $||\bm{\theta}^{est}-\bm{\theta}||_{2}$, is shown in (a) of \figref{figure:musashi-exp}.
  It can be seen that $||\bm{\theta}^{est}-\bm{\theta}||_{2}$ decreases gradually as the online update is executed.
  The exponential approximation of the results of the 10-minute experiment shows that the joint angle estimation error decreases from 0.324 rad to 0.154 rad in 10 minutes, which is about half of the original value.

  We evaluate the state estimation using GeMuCo.
  We stopped the online updater after it was performed, and evaluated the joint angle estimation error before and after the update.
  While the robot grasps a heavy object and then stops the function of one muscle, the joint angle estimation error and the anomaly $d$ are measured.
  The experimental results are shown in (b) of \figref{figure:musashi-exp}.
  When the joints are moved randomly, the joint angle estimation error drops significantly from 0.414 rad to 0.186 rad on average after the online update.
  In addition, since the space of muscle tension is also learned during the online update, the joint angle estimation error does not increase significantly even when a heavy object is grasped.
  % Finally, we observe the transition of the degree $d$ for anomaly detection.
  After the online update, $d$ rises sharply when stopping the function of one muscle, indicating that the robot can detect an anomaly.
  On the other hand, $d$ does not rise much before the online update.
  This is because the muscle tension is determined randomly in the space as a whole during the initial training, so that the sensor values can be reconstructed even with infeasible muscle tension.

  We evaluate the muscle length-based joint position control.
  The loss function is set as follows,
  \begin{align}
    L = &||\bm{\theta}^{pred}-\bm{\theta}^{ref}||_{2} + ||\bm{f}^{pred}||_{2}\nonumber\\ &+ 0.01||\bm{\tau}_{ext}+\bm{G}^{T}(\bm{\theta}^{ref}, \bm{f}^{pred})\bm{f}^{pred}||_{2}
  \end{align}
  where $\{\bm{\theta}, \bm{f}\}^{pred}$ is the predicted value from the network, $\bm{\theta}^{ref}$ is the target joint angle, $\bm{G}$ is the muscle Jacobian obtained by differentiating the network, and $\bm{\tau}_{ext}$ is the desired joint torque (mainly gravity compensation torque).
  First, the target joint angle $\bm{\theta}^{ref}$ to be evaluated is randomly determined at five points.
  The motion from a random joint angle $\bm{\theta}_{rand}$ to $\bm{\theta}^{ref}$ is performed five times while changing $\bm{\theta}_{rand}$, and the average and variance of the control error $||\bm{\theta}^{ref}-\bm{\theta}||_{2}$ are evaluated.
  The results when using the model before and after the online update are shown in (c) of \figref{figure:musashi-exp}.
  It can be seen that the control after online update is more accurate in realizing the joint angle than that before the online update.

  We evaluate the simulation constructed by GeMuCo.
  The loss function for the simulator is set as follows,
  \begin{align}
    L = &||\bm{l}^{pred}-\bm{l}^{send}||_{2} + 0.1||\bm{f}^{pred}||_{2}\nonumber\\ &+ 0.001||\bm{\tau}_{ext}+G^{T}(\bm{\theta}^{pred}, \bm{f}^{pred})\bm{f}^{pred}||_{2} \label{eq:musashi-sim1}
  \end{align}
  where $\bm{l}^{send}$ refers to the muscle length commanded to the simulation and the calculated $\{\bm{\theta}, \bm{f}\}^{pred}$ is used as the simulated value $\{\bm{\theta}, \bm{f}\}^{sim}$.
  % We evaluate the difference between the actual robot motion and the simulated robot motion before and after the online update.
  The comparison of sensor values during the simulation before and after online update and the actual robot motion is shown in (d) of \figref{figure:musashi-exp}.
  The higher the value of muscle tension, the redder the color of the muscle is, and the smaller the value of muscle tension, the greener the color is.
  It can be seen that the simulated muscle tension and joint angle are closer to the sensor values of the actual robot after the online update than before.
  Transitions of $||\bm{\theta}^{sim}-\bm{\theta}||_{2}$ and $||\bm{f}^{sim}-\bm{f}||_{2}$ before and after the online update are shown in (e) of \figref{figure:musashi-exp}.
  The average error between the actual sensor values and the simulated sensor values changes from 0.335 rad to 0.162 rad for the joint angles and from 104.8 N to 92.2 N for muscle tensions, which are smaller after the online update than before.
  This indicates that it is possible to make the behavior of the simulation closer to that of the actual robot by online update of GeMuCo.
  In addition, the simulation behavior can be modified by changing the loss function as shown in (f) of \figref{figure:musashi-exp}.
  This is the behavior when the elbow pitch angle $\bm{\theta}_{E-p}$ is bent at 90 degrees, -50 N is applied in the $z$ direction, 50 N in the $y$ direction, and then the shoulder pitch angle $\theta_{S-p}$ is forced to 30 deg.
  The loss function of \equref{eq:musashi-sim1} is sufficient when specifying the force, but the loss function is set as the following when specifying the posture $\bm{\theta}_{fix}$ forcibly.
  \begin{align}
    L = &||\bm{l}^{pred}-\bm{l}^{send}||_{2} + 0.1||\bm{f}^{pred}||_{2}\nonumber\\ &+ ||\bm{\theta}^{pred}-\bm{\theta}_{fix}||_{2} \label{eq:musashi-sim2}
  \end{align}
  By changing the applied force or $\bm{\theta}_{fix}$, it is possible to check the corresponding changes in joint angle and muscle tension.
}%
{%
  まず, 幾何モデルを用いたネットワークの初期学習を行う.
  ランダムな関節角度と筋張力から筋長を求め, 10000のデータからネットワークを初期化する.
  次に, 実機データに基づくオンライン学習を行う.
  ここでは, ランダムな関節角度と筋張力からロボットを動かすことを繰り返し, 学習を行う.
  この際の関節角度推定値$\bm{\theta}^{est}$と関節モジュールのセンサ値$\bm{\theta}$の差分$||\bm{\theta}^{est}-\bm{\theta}||_{2}$の遷移を\figref{figure:musashi-exp}の(a)に示す.
  オンライン学習を実行するにつれ, $||\bm{\theta}^{est}-\bm{\theta}||_{2}$が徐々に下がっていくことがわかる.
  10分間の実験を行った結果を指数関数近似した曲線から, 関節角度推定誤差は10分間で0.324 radから0.154 radと, 約半分程度まで下がっていることがわかる.

  状態推定について実機実験を行い評価する.
  オンライン学習を行った後にそれを止め, 学習する前とした後における関節角度推定値の追従の差を評価した.
  実験の途中で, 重量物体の把持, また, 筋一本の機能を停止することを行い, その際の関節角度推定値の追従, また, 異常度$d$の測定を行う.
  本実験中は一切のオンライン学習は行っていない.
  実験結果を\figref{figure:musashi-exp}の(b)に示す.
  ランダムに関節を動かした際には, 学習前と後で, 関節角度推定誤差が, 平均で0.414 radから0.186 radと, 大きく下がっていることがわかる.
  また, オンライン学習の際に筋張力の空間も効率的に学習しているため, 重量物体を把持した際も大きく関節角度推定誤差が上がることはなかった.
  最後に, 異常検知度$d$の推移を観察する.
  学習後は, 一本の筋の機能を停止することで, $d$が急上昇しており, 異常を検知可能なことがわかった.
  一方で, 学習前では$d$はあまり上昇しない.
  これは, 初期学習の際には筋張力を全空間でランダムに決めているため, 実行不可能な筋張力においても同じようにセンサ値を復元可能であり, $d$が下がるためである.

  筋長による関節位置制御について実機実験を行い評価する.
  この際の損失関数は以下のように設定した.
  \begin{align}
    L = &||\bm{\theta}^{pred}-\bm{\theta}^{ref}||_{2} + ||\bm{f}^{pred}||_{2}\nonumber\\ &+ 0.01||\bm{\tau}_{ext}+\bm{G}^{T}(\bm{\theta}^{ref}, \bm{f}^{pred})\bm{f}^{pred}||_{2}
  \end{align}
  ここで, $\{\bm{\theta}, \bm{f}\}^{pred}$はネットワークの予測値, $\bm{\theta}^{ref}$は指令関節角度, $\bm{G}$はネットワークを微分することで得られる筋長ヤコビアン, $\bm{\tau}_{ext}$は発揮したい関節トルク(基本的には重力補償トルク)である.
  まず, 評価を行う関節角度$\bm{\theta}^{ref}$を5点ランダムに決定する.
  あるランダムな関節角度$\bm{\theta}_{rand}$から$\bm{\theta}^{ref}$に動作することを$\bm{\theta}_{rand}$を変えながら5回行い, そのときの制御誤差$||\bm{\theta}^{ref}-\bm{\theta}||_{2}$の平均と分散を評価する.
  学習前のモデルを使った場合と学習後のモデルを使った場合の結果を\figref{figure:musashi-exp}の(c)に示す.
  学習前よりも, 学習後の制御の方が関節角度の実現精度が高いことがわかる.

  シミュレーション構築に関する実験を行う.
  この際の損失関数は以下のように設定した.
  \begin{align}
    L = &||\bm{l}^{pred}-\bm{l}^{send}||_{2} + 0.1||\bm{f}^{pred}||_{2}\nonumber\\ &+ 0.001||\bm{\tau}_{ext}+G^{T}(\bm{\theta}^{pred}, \bm{f}^{pred})\bm{f}^{pred}||_{2} \label{eq:musashi-sim1}
  \end{align}
  ここで, $\bm{l}^{send}$はシミュレーションに送られた指令筋長を指し, 最終的に計算された$\{\bm{\theta}, \bm{f}\}^{pred}$をシミュレーションの値$\{\bm{\theta}, \bm{f}\}^{sim}$とする.
  オンライン学習前と学習後において, 実機の動きとシミュレーション上のロボットの動きの差異を評価する.
  学習前のシミュレーション・学習後のシミュレーション・実機動作の際におけるセンサ値の比較を\figref{figure:musashi-exp}の(d)に示す.
  筋張力の値が高いほど筋の色が赤く, 小さいほど色が緑色になるようにしている.
  学習前に比べ, 学習後の方が実機センサ値に筋張力・関節角度が近づいていることがわかる.
  学習前と学習後における$||\bm{\theta}^{sim}-\bm{\theta}||_{2}$と$||\bm{f}^{sim}-\bm{f}||_{2}$の遷移を\figref{figure:musashi-exp}の(e)に示す.
  学習前に比べ学習後の方が, 関節角度については平均で0.335 radから0.162 rad, 筋張力については平均で104.8 Nから92.2 Nと, 実機センサ値とシミュレーションセンサ値の誤差が小さくなっていることがわかる.
  これにより, シミュレーションの挙動を学習により実機に近づけていくことが可能であることがわかった.
  また, 損失関数を変えることでシミュレーションの挙動を変化させることができる様子を\figref{figure:musashi-exp}の(f)に示す.
  これは, 肘を90度に曲げ, $z$方向に-50 N, $y$方向に50 Nを順にかけ, その後$\theta_{S-p}$を無理やり30 degにした際の挙動である.
  力を指定する際は\equref{eq:musashi-sim1}で可能だが, 姿勢$\bm{\theta}_{fix}$を指定する際は以下の\equref{eq:musashi-sim2}のように損失関数を設定している.
  \begin{align}
    L = &||\bm{l}^{pred}-\bm{l}^{send}||_{2} + 0.1||\bm{f}^{pred}||_{2}\nonumber\\ &+ ||\bm{\theta}^{pred}-\bm{\theta}_{fix}||_{2} \label{eq:musashi-sim2}
  \end{align}
  かける力を変えたり, $\bm{\theta}_{fix}$を与えたりすることで, それに応じた関節角度・筋張力を確認することが可能である.
}%

\subsection{Full-Body Tool Manipulation Learning for Low-Rigidity Humanoids}
\switchlanguage%
{%
  In this experiment, a low-rigidity humanoid manipulates tools while maintaining balance with its body.
  Due to its low rigidity, the deflection of the body changes depending on the length and weight of the tool, and it is necessary to detect such changes while manipulating the tool.
}%
{%
  本実験では低剛性なヒューマノイドが全身を使ってバランスを取りながら道具操作を行う.
  低剛性ゆえに道具の長さや重量によって身体の撓みが変化するため, その変化を検知しながら道具操作を行う必要がある.
}%

\begin{figure*}[t]
  \centering
  \includegraphics[width=1.95\columnwidth]{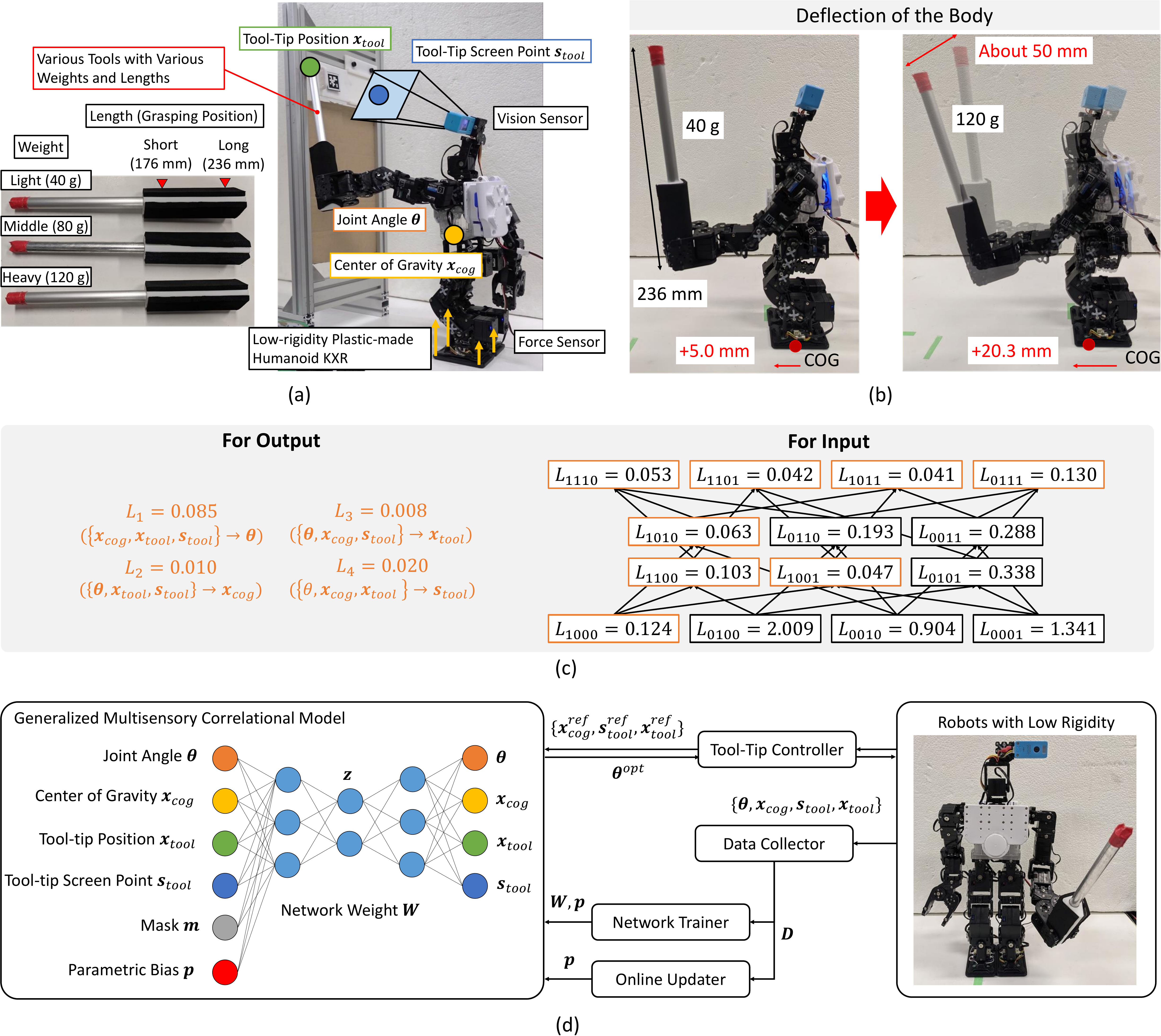}
  \caption{(a) Full-body tool manipulation learning for low-rigidity humanoids with sensors of joint angle, center of gravity, tool-tip screen point, and tool-tip position. We prepared six types of tool states with various weights and lengths. (b) Due to the low rigidity of the hardware, the deflection of the body changes depending on the length and weight of the tool. (c) Automatic determination of the network structure for full-body tool manipulation learning. (d) The system configuration for full-body tool manipulation learning.}
  \label{figure:kxr-overview}
\end{figure*}

% \begin{figure}[t]
%   \centering
%   \includegraphics[width=0.95\columnwidth]{figs/kxr-overview2}
%   \caption{The change in tool-tip position and center of gravity when handling tools with different weights.}
%   \label{figure:kxr-overview2}
% \end{figure}

\subsubsection{\textbf{Experimental Setup}}
\switchlanguage%
{%
  In this experiment, KXR, a low-rigidity plastic-made humanoid, is used.
  We set $\bm{x}=\{\bm{\theta}, \bm{x}_{cog}, \bm{x}_{tool}, \bm{s}_{tool}\}$.
  Here $\bm{\theta} = \begin{pmatrix}\theta_{S-p} & \theta_{S-y} & \theta_{E-p} & \theta_{A-p}\end{pmatrix}^{T}$ is the commanded joint angle ($S$ for the shoulder, $E$ for the elbow, $A$ for the ankle, and $py$ for the pitch and yaw, respectively).
  $\bm{x}_{cog}=\begin{pmatrix}x_{cog} & y_{cog}\end{pmatrix}^{T}$ is the position of the center of gravity calculated from the single-axis force sensors placed at the four corners of each sole (since the position of the foot is not changed during tool manipulation, the feet are assumed to be aligned. $z$ axis is ignored).
  $\bm{x}_{tool}$ denotes the tool-tip position in the 3D space recognized by AR marker attached to the tool-tip.
  $\bm{s}_{tool}$ denotes the tool-tip position in the 2D space in the image recognized by color extraction.
  If an AR maker is attached to the tool-tip, $\{\bm{x}_{tool}, \bm{s}_{tool}\}$ is obtained, while only $\bm{s}_{tool}$ is obtained if there is no AR marker and only color extraction is executed.
  We attach the AR marker when collecting the training data, but this is removed when using the trained network because the AR marker is obstructive to tool manipulation.

  In the experiments, we use a simulation and the actual robot of KXR.
  For both cases, we prepare tools with three different weights (Light: 40 g, Middle: 80 g, Heavy: 120 g) as shown in (a) of \figref{figure:kxr-overview}.
  In addition, we define two lengths (Short: 176 mm, Long: 236 mm) according to the grasped position of the tools, and handle a total of six tool states, which are combinations of these weights and lengths.
  (b) of \figref{figure:kxr-overview} shows the changes in the tool-tip position and the center-of-gravity position when the actual KXR holds tools with different weights.
  For the same tool length, the tool-tip position differs by about 50 mm and the center of gravity by about 15 mm when holding a 40 g tool and a 120 g tool, due to the low rigidity of the hardware.
  Similarly, if the length of the tool changes, the tool-tip positions $\bm{x}_{tool}$ and $\bm{s}_{tool}$ change, and at the same time, the center of gravity $\bm{x}_{cog}$ also changes.
  In order to simply reproduce the deflection of the actual robot in the simulation, each joint is simulated to be deflected by $30\bm{\tau}$ [deg] for the joint torque $\bm{\tau}$ [Nm] applied to the body.

  Note that $\bm{p}$ is assumed to be two-dimensional in each experiment.
}%
{%
  本研究では低剛性な樹脂製ヒューマノイドKXRを実験に用いる.
  本研究では$\bm{x}=\{\bm{\theta}, \bm{x}_{cog}, \bm{x}_{tool}, \bm{s}_{tool}\}$としている.
  ここで, $\bm{\theta}=\begin{pmatrix}\theta_{S-p} & \theta_{S-y} & \theta_{E-p} & \theta_{A-p}\end{pmatrix}^{T}$は肩のpitchとyawの角度(S-p, S-y), 肘のpitchの角度(E-p), そして足首のpitchの角度(A-p)の4つの角度の指令値である.
  $\bm{x}_{cog}=\begin{pmatrix}x_{cog} & y_{cog}\end{pmatrix}^{T}$はそれぞれの足裏の四隅に配置された1軸の力センサから算出した重心位置(道具操作中は足平の位置は変更しないため, 足は揃っているという仮定を置いている, また, z方向は無視する).
  $\bm{x}_{tool}$は道具先端位置をARマーカにより認識した際の, その3次元空間上での位置を表す.
  $\bm{s}_{tool}$は道具先端位置を色認識等によって認識した際の, その画像上の2次元の位置を表す.
  実機においては, 学習データ取得時は道具先端にARマーカをつけることで$\{\bm{x}_{tool}, \bm{s}_{tool}\}$を取得するが, 動作実行時はARマーカが道具操作に邪魔であるため, 道具先端に色を付けて$\bm{s}_{tool}$のみを取得している.

  本研究の実験では, KXRのシミュレーションと実機を用いて実験を行う.
  両者について, \figref{figure:kxr-overview}の(a)に示すように, 3種類の重さ(Light: 40 g, Middle: 80 g, Heavy: 120 g)の道具を用意する.
  また, それらの道具の持つ位置に応じて, 2種類の長さ(Short: 全長176 mm, Long: 全長236mm)を定義し, これらの組み合わせである計6種類の道具状態を扱う.
  \figref{figure:kxr-overview}の(b)に, KXR実機が異なる重さの道具を持った際の道具先端位置と重心位置の変化を示す.
  同じ長さの道具でも, ハードウェアが低剛性であるがゆえに, 40 gの道具と120 gの道具を持った時では道具先端位置は約50 mm, 重心位置は約15mmも異なる.
  同様に, 道具の長さが変わればその道具先端位置$\bm{x}_{tool}$や$\bm{s}_{tool}$が変化すると同時に, 重心位置$\bm{x}_{cog}$も変化する.
  実機における撓みをシミュレーション上で簡単に再現するために, 身体にかかる関節トルク$\bm{\tau}$ [Nm]に対して, $30\bm{\tau}$ [deg]だけそれぞれの関節を撓ませるようにシミュレーションを行っている.

  なお, それぞれの実験で$\bm{p}$は2次元とした.
}%

\begin{figure*}[t]
  \centering
  \includegraphics[width=1.7\columnwidth]{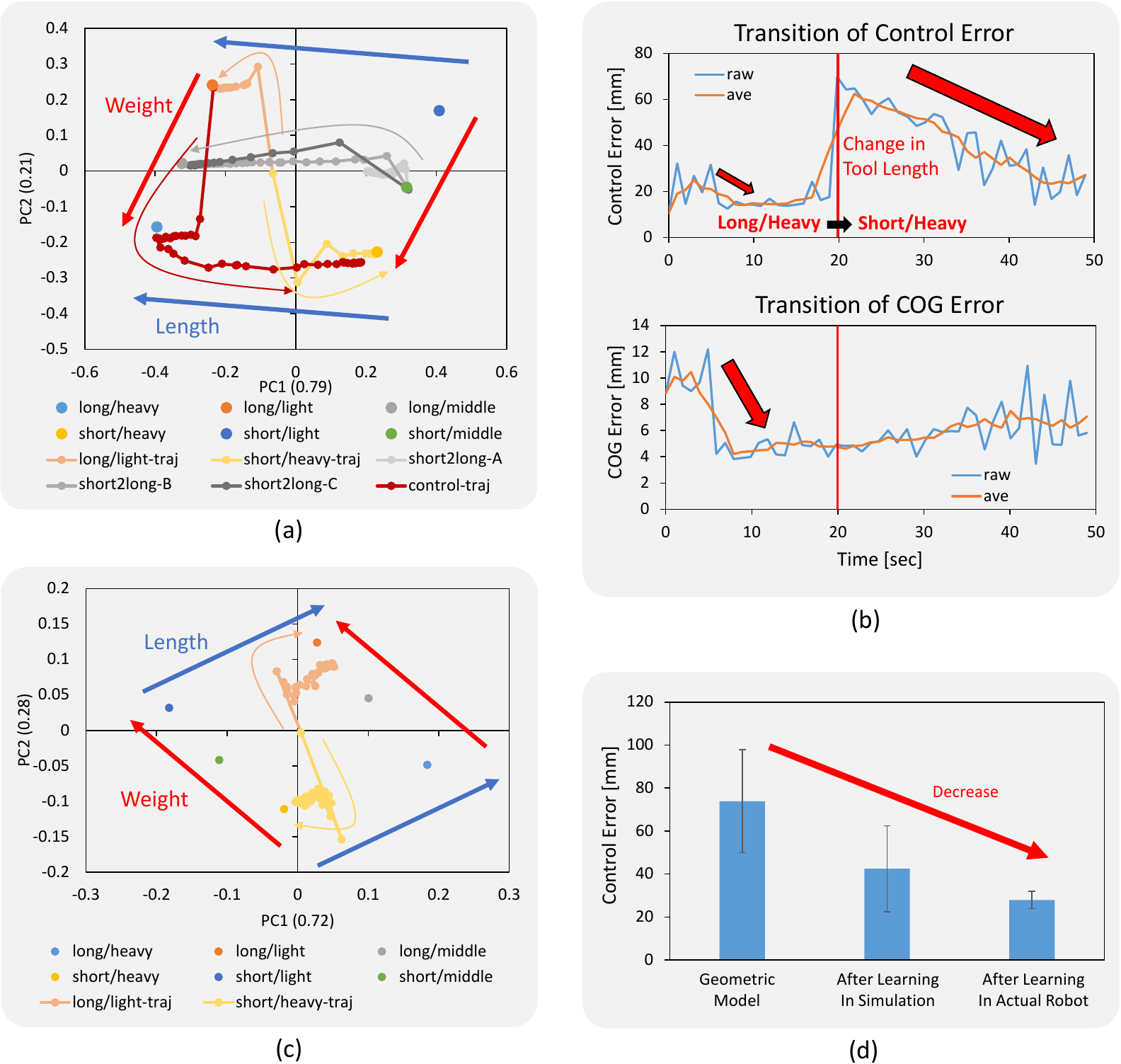}
  \caption{(a) The trained parametric bias and its trajectory during online update of GeMuCo in the simulation experiment. (b) The transition of control and center-of-gravity errors in the simulation experiment. (c) The trained parametric bias and its trajectory during online update of GeMuCo in the actual robot experiment. (d) Comparison of control errors among when using a geometric model, using GeMuCo after training in simulation, and using GeMuCo after training in the actual robot \cite{kawaharazuka2024kxr}.}
  \label{figure:kxr-exp}
\end{figure*}

\subsubsection{\textbf{Network Structure}}
\switchlanguage%
{%
  In the simulation, we randomly move the joint angle, and obtain 500 data points per one tool state, amounting to 3000 data points in total.
  The automatic determination of the network structure based on the obtained data is shown in (c) \figref{figure:kxr-overview}.
  First, $L_{i}$ is computed and all values in $\bm{x}$ are used as $\bm{x}_{out}$, since all are less than $C^{out}_{thre}$.
  Next, $L_{m}$ is computed for each mask $\bm{m}$, and the feasible mask set is determined to be $\mathcal{M}=\{\begin{pmatrix}1&1&1&0\end{pmatrix}^{T}$, $\begin{pmatrix}1&1&0&1\end{pmatrix}^{T}$, $\begin{pmatrix}1&0&1&1\end{pmatrix}^{T}$, $\begin{pmatrix}0&1&1&1\end{pmatrix}^{T}$, $\begin{pmatrix}1&0&1&0\end{pmatrix}^{T}$, $\begin{pmatrix}1&1&0&0\end{pmatrix}^{T}$, $\begin{pmatrix}1&0&0&1\end{pmatrix}^{T}$, $\begin{pmatrix}1&0&0&0\end{pmatrix}^{T}\}$.
  All values of $\bm{x}$ are used as $\bm{x}_{in}$.
  The network $\{\bm{\theta}, \bm{x}_{cog}, \bm{x}_{tool}, \bm{s}_{tool}\}\rightarrow\{\bm{\theta}, \bm{x}_{cog}, \bm{x}_{tool}, \bm{s}_{tool}\}$ and the system configuration shown in (d) of \figref{figure:kxr-overview} are automatically constructed.
}%
{%
  シミュレーションにおいて関節角度をランダムに動かし, 一つの道具状態に対して500, 全体で3000のデータを取得した.
  得られたデータに基づくネットワーク構造の自動決定について\figref{figure:kxr-inout}の(c)に示す.
  まず, $L_{i}$を計算し, そのどれもが$C^{out}_{thre}$以下であるため, $\bm{x}$における全ての値を$\bm{x}_{out}$として用いることに決定した.
  次に, それぞれのマスク$\bm{m}$について$L_{m}$を計算し, 実行可能なマスク集合$\mathcal{M}=\{\begin{pmatrix}1&1&1&0\end{pmatrix}^{T}$, $\begin{pmatrix}1&1&0&1\end{pmatrix}^{T}$, $\begin{pmatrix}1&0&1&1\end{pmatrix}^{T}$, $\begin{pmatrix}0&1&1&1\end{pmatrix}^{T}$, $\begin{pmatrix}1&0&1&0\end{pmatrix}^{T}$, $\begin{pmatrix}1&1&0&0\end{pmatrix}^{T}$, $\begin{pmatrix}1&0&0&1\end{pmatrix}^{T}$, $\begin{pmatrix}1&0&0&0\end{pmatrix}^{T}\}$が決定され, これに必要な$\bm{x}_{in}$として, $\bm{x}$の全ての値が設定された.
    よって, $\{\bm{\theta}, \bm{x}_{cog}, \bm{x}_{tool}, \bm{s}_{tool}\}\rightarrow\{\bm{\theta}, \bm{x}_{cog}, \bm{x}_{tool}, \bm{s}_{tool}\}$というネットワークが構築され, \figref{figure:kxr-system}の(d)に示すようなシステム構成が自動構築される.
}%

\subsubsection{\textbf{Simulation Experiment}}
\switchlanguage%
{%
  First, we randomly move the joint angle in the simulation, obtain 500 data per one tool state, amounting to 3000 data in total, and train the network based on these data.
  The trained parametric bias $p_k$ is represented in two-dimensional space through PCA as shown in (a) of \figref{figure:kxr-exp}.
  We can see that each parametric bias is neatly self-organized along the weight and length of the tool.

  Next, we show that the online update of parametric bias can make possible to accurately recognize the current tool state.
  We set the tools to the Long/Light and Short/Heavy states and performed the online update of parametric bias while commanding random joint angles.
  The transitions of the parametric bias are shown in (a) of \figref{figure:kxr-exp} (``long/light-traj'' and ``short/heavy-traj'').
  It can be seen that the current PB values are gradually approaching the Long/Light or Short/Heavy PB values obtained at  training phase.

  Next, we consider how the online update is affected by the different types of sensors used.
  If an AR maker is attached to the tool-tip, $\{\bm{x}_{tool}, \bm{s}_{tool}\}$ is obtained, while only $\bm{s}_{tool}$ is obtained if there is no AR marker and only color extraction is executed.
  Moreover, when the tool-tip is not within the sight of vision, we can only obtain $\{\bm{\theta},\bm{x}_{cog}\}$.
  Therefore, in this experiment, there are three cases for the obtained data: A $\{\bm{\theta}, \bm{x}_{cog}\}$, B $\{\bm{\theta}, \bm{x}_{cog}, \bm{s}_{tool}\}$, C $\{\bm{\theta}, \bm{x}_{cog}, \bm {x}_{tool}, \bm{s}_{tool}\}$, and we examine how the PB values transition in these three cases.
  Here, we start from the PB for Short/Middle obtained at the training phase and examine the case of grasping Long/Middle tools.
  The results are shown in (a) of \figref{figure:kxr-exp} as ``short2long-A'', ``short2long-B'', and ``short2long-C''.
  Case C has the same sensor type as the aforementioned ``long/light-traj'' and ``short/heavy-traj'', and the PB transitions quickly in 15 steps, indicating that the tool state can be accurately recognized.
  In case B, the transition speed is slower than that of case C, and it takes 35 steps to accurately recognize the tool state.
  In case A, although PB is moving in the correct direction, it does not reach the correct value even after 35 steps of online update as in case B.

  Finally, we conduct a control experiment including online update of parametric bias.
  We start from the state where the tool state is correctly recognized as Long/Light, and change the actual tool state in the order of Long/Heavy and Short/Heavy.
  The random target tool-tip position $\bm{x}^{ref}_{tool}$ and the constant target center of gravity $\bm{x}^{ref}_{cog}=\begin{pmatrix}0&0\end{pmatrix}^{T}$ (coordinates of the center of the foot) are given within the specified range, and GeMuCo is used to follow these target values.
  The loss function in this case is as follows.
  \begin{align}
    L = ||\bm{x}^{pred}_{tool}-\bm{x}^{ref}_{tool}||_{2} + 0.01||\bm{x}^{pred}_{cog}-\bm{x}^{ref}_{cog}||_{2}
  \end{align}
  The transition of parametric bias is shown in ``control-traj'' of (a) of \figref{figure:kxr-exp}.
  It can be seen that the parametric bias transitions from Long/Light to Long/Heavy to Short/Heavy, in order.
  The control error $||\bm{x}^{ref}_{tool}-\bm{x}_{tool}||_{2}$ and the center-of-gravity error $||\bm{x}^{ref}_{cog}-\bm{x}_{cog}||_{2}$ (raw) and each average of 5 steps (ave) are shown in (b) of \figref{figure:kxr-exp}.
  We can see that in the Long/Heavy condition, the control error is slightly decreased, and the center-of-gravity position error is significantly decreased by updating the PB value.
  It is because the tool state is changed from Long/Light to Long/Heavy and the weight of the tool changes significantly while the length of the tool remains the same.
  When the tool state is shifted to the Short/Heavy condition, the control error changes significantly, but the center-of-gravity error does not change significantly.
  Similarly, the control error decreases significantly by online update of parametric bias.
}%
{%
  先と同様にシミュレーションにおいて関節角度をランダムに動かし, 一つの道具状態に対して500, 全体で3000のデータを取得し, これを元にネットワークを学習させる.
  このときに得られたParametric Bias $p_k$をPCAを通して2次元空間に表現した図を\figref{figure:kxr-exp}の(a)に示す.
  それぞれのParametric Biasが, 道具の重さと長さの軸に沿って綺麗に自己組織化していることがわかる.

  次に, このParametric Biasをオンライン学習することで, 現在の道具状態を正確に認識できることを示す.
  道具をLong/Lightの状態, Short/Heavyの状態に設定し, ランダムな関節角度を送りながらParametric Biasのオンライン学習を実行した.
  その際のParametric Biasの遷移を\figref{figure:kxr-exp}の(a)に示す(long/light-trajとshort/heavy-traj).
  現在のPB値が, 訓練時のLong/Light, またはShort/HeavyのPB値へと徐々に近づいていることがわかる.

  次に, 動作時に得られるセンサの種類の違いにより, どうオンライン学習が変化するかを考察する.
  道具先端にARマーカがついていれば$\{\bm{x}_{tool}, \bm{s}_{tool}\}$が得られるが, ARマーカなしで色認識のみの場合は$\bm{s}_{tool}$しか得られない.
  また, 視覚に道具先端が入らない場合は, $\{\bm{\theta}, \bm{x}_{cog}\}$の値しか得られない.
  そこで, 本研究では得られるデータがA: $\{\bm{\theta}, \bm{x}_{cog}\}$, B: $\{\bm{\theta}, \bm{x}_{cog}, \bm{s}_{tool}\}$, C: $\{\bm{\theta}, \bm{x}_{cog}, \bm{x}_{tool}, \bm{s}_{tool}\}$の3種類の場合で, どのようにPB値が遷移するかを検証する.
  ここでは, Short/MiddleにおけるPBからスタートし, Long/Middleの道具を扱った場合について検証する.
  その結果を同様にshort2long-A, short2long-B, short2long-Cとして\figref{figure:kxr-exp}の(a)に示す.
  Cは前述のlong/light-trajやshort/heavy-trajとセンサの種類が同じであり, 15ステップで素早くPBが遷移し, 正確に道具状態を把握できている.
  BはCよりも遷移のスピードが落ち, 正確に道具状態を把握するのに35ステップの時間を要している.
  Aは正しい方向にPB値自体は進んでいるものの, Bと同じ35ステップのオンライン学習を実行しても正しくPBを認識するところまでには至らなかった.

  最後に, Parametric Biasのオンライン学習を含んだ制御実験を行う.
  道具状態をLong/Lightとして正しく認識している状態から始め, 実際の道具状態をLong/Heavy, Short/Heavyの順で変更する.
  % なお, ここでは道具変化にすぐ対応できるように, $N^{online}_{max}=5$とした.
  指定した範囲内のランダムな$\bm{x}^{ref}_{tool}$と指令重心位置$\bm{x}^{ref}_{cog}=\begin{pmatrix}0&0\end{pmatrix}^{T}$ (足平中心座標)を与え, これに追従するよう, GeMuCoを使った動作制御を実行する.
  この際の損失関数は以下である.
  \begin{align}
    L = ||\bm{x}^{pred}_{tool}-\bm{x}^{ref}_{tool}||_{2} + 0.01||\bm{x}^{pred}_{cog}-\bm{x}^{ref}_{cog}||_{2}
  \end{align}
  このときのParametric Biasの遷移を\figref{figure:kxr-exp}の(a)のcontrol-trajに示す.
  Long/LightからLong/Heavy, Short/Heavyの順にParametric Biasが遷移していることがわかる.
  また, このときの制御誤差$||\bm{x}^{ref}_{tool}-\bm{x}_{tool}||_{2}$と重心位置誤差$||\bm{x}^{ref}_{cog}-\bm{x}_{cog}||_{2}$ (raw), それぞれの5ステップ分の平均 (ave)を\figref{figure:kxr-exp}の(b)に示す.
  まず, Long/Heavyの状態では, PB値が更新されることで, 制御誤差が少し, また, 重心位置誤差が大きく減少していることがわかる.
  PB値がLong/LightからLong/Heavyに変化し, 道具の長さは変わらないが重さが大きく変化したため, 制御誤差に比べて重心位置誤差が大きく減少したのだと考えられる.
  その後, Short/Heavyの状態に移行した際には, 大きく制御誤差が変化するが, 重心位置誤差に大きな変化はない.
  同様にPB値のオンライン学習により, 制御誤差が大きく減少していく.
  一方, 重心位置誤差については大きく変化していない.
}%

\subsubsection{\textbf{Actual Robot Experiment}}
\switchlanguage%
{%
  First, (1) data collection using a GUI (60 data points) and (2) data collection using random joint angles (20 data points) are performed for each tool, and a total of about 480 data points are collected.
  In (1), data is collected when a human directly specifies $\bm{\theta}$ using a GUI.
  In (2), $\bm{\theta}$ is specified randomly in the simulation, and is commanded to the actual robot when the center of gravity is within the support area and the tool-tip position is visible from the camera.
  By strengthening the constraint on the position of the center of gravity, it is possible to collect data to the extent that the robot does not fall over even if the simulation and the actual robot are different.
  Since the data obtained from the actual robot is limited, the model generated from the aforementioned simulation is fine-tuned.
  The arrangement of parametric bias obtained in this process is shown in (c) of \figref{figure:kxr-exp}.
  We can see that each parametric bias is neatly self-organized along the weight and length of the tool.

  Next, we show that online update of this parametric bias can make it possible to accurately recognize the current tool state.
  We set the tools to Long/Light and Short/Heavy states, and perform online update of parametric bias while commanding random joint angles.
  The transition of the parametric bias is shown in (c) of \figref{figure:kxr-exp} (``long/light-traj'' and ``short/heavy-traj'').
  It can be seen that the current PB values are gradually approaching the Long/Light and Short/Heavy PB values obtained at the training.

  Finally, we evaluate the control error.
  (d) of \figref{figure:kxr-exp} shows the comparison of the control errors at the Long/Middle tool state in the case of solving the full-body inverse kinematics using the geometric model, in the case of training GeMuCo from the simulation data including joint deflection, and in the case of fine-tuning GeMuCo using the actual robot sensor data.
  Note that the parametric bias when using GeMuCo is that of Long/Middle obtained at the training.
  It can be seen that the geometric model has the largest error, and the control error becomes smaller after training by simulation including joint deflection, and even smaller after training by the actual robot data.
}%
{%
  実機において(1) GUIを用いたデータ取得(60データ)と(2) 関節角度のランダム指定を用いたデータ取得(20データ)をそれぞれの道具に対して行い, 全部で約480のデータを取得した.
  (1)はGUIを用いて人間が$\bm{\theta}$を直接指定し, そのときのデータを収集していく方法である.
  (2)はシミュレーションにおいて$\bm{\theta}$をランダムに指定し, このとき重心が支持領域内に含まれ, かつ道具先端位置がカメラから見える場合の$\bm{\theta}$を実機に送って動かす方法である.
  重心位置に関する制約を強くすることで, たわみによりシミュレーションと実機が異なっても, 倒れない範囲でデータを収集することが可能である.
  実機で得られるデータは少数なため, 前述のシミュレーションから生成されたモデルをFine-Tuningする.
  この際に得られたParametric Biasの配置を\figref{figure:kxr-exp}の(c)に示す.
  それぞれのParametric Biasが, 道具の重さと長さの軸に沿って綺麗に自己組織化していることがわかる.

  次に, このParametric Biasをオンライン学習することで, 現在の道具状態を正確に認識できることを示す.
  道具をLong/Lightの状態, Short/Heavyの状態に設定し, ランダムな関節角度を送りながらParametric Biasのオンライン学習を実行した.
  その際のParametric Biasの遷移を同様に\figref{figure:kxr-exp}の(c)に示す(long/light-trajとshort/heavy-traj).
  現在のPB値が, 訓練時のLong/Light, またはShort/HeavyのPB値へと徐々に近づいていることがわかる.

  最後に, 制御誤差に関する評価を行う.
  Long/Middleの道具状態において, 幾何モデルを用いて全身逆運動学を解いた場合, 関節のたわみを含むシミュレーションデータからGeMuCoを学習させた場合, これを実機においてFine-Tuningした場合について制御誤差を比較した結果を\figref{figure:kxr-exp}の(d)に示す.
  なお, GeMuCoを用いる際のParametric Biasは訓練時のLong/Middleの値である.
  幾何モデルが最も誤差が大きく, 関節のたわみを含むシミュレーションによる学習後, 実機による学習後の順で制御誤差が小さくなっていることがわかる.
}%

\section{Discussion and Limitations} \label{sec:discussion}

\subsection{Discussion}
\switchlanguage%
{%
  % We summarize the experimental results obtained so far.
  First, in the tool-tip control experiment of PR2, we handled a very simple network configuration $\bm{\theta}\rightarrow\bm{x}_{tool}$.
  It is automatically detected that $\bm{\theta}\rightarrow\bm{x}_{tool}$ is computable, but the reverse is not.
  By collecting data in various grasping states, the information on the changes of these grasping states is embedded in parametric bias, and by updating it online, the current grasping state can always be recognized.
  In addition, the tool-tip position can be estimated and controlled based on the forward and backward propagation and gradient descent of the network, and these become more accurate when the grasping state is correctly recognized.
  Online update of parametric bias can maintain high generalization performance, but updating the network weight $\bm{W}$ is prone to loss of generalization performance.
  Similarly, in the actual robot experiment with a flexible tool, online update of parametric bias and tool-tip control are possible, demonstrating the effectiveness of this system.

  Next, in the body control experiment of the musculoskeletal humanoid Musashi, we handled a complex network configuration $(\bm{\theta}, \bm{f}, \bm{l})\rightarrow(\bm{\theta}, \bm{f}, \bm{l})$.
  There is a very complex relationship between $\{\bm{\theta}, \bm{f}, \bm{l}\}$ in the musculoskeletal system, and in general there are three relationships: $\{\bm{\theta}, \bm{f}\}\rightarrow\bm{l}$, $\{\bm{f}, \bm{l}\}\rightarrow\bm{\theta}$, and $\{\bm{\theta}, \bm{l}\}\rightarrow\bm{f}$.
  By expressing these relationships in a single network, state estimation, control, and simulation become possible.
  The network is initialized using data obtained from the geometric model and updated online using the actual robot sensor data.
  Unlike the aforementioned experiment, this is an example of updating $\bm{W}$ directly because it does not include parametric bias, but overfitting can be avoided by collecting data with random joint angles and muscle tensions.
  By updating the network with actual robot sensor data, the accuracy of state estimation, control, and simulation is improved.
  In terms of simulation, the proposed method can simulate various situations depending on the definition of the loss function, showing the versatility of the proposed method.

  Finally, in the full-body tool manipulation experiment of the low-rigidity humanoid KXR, we handled an even more complex network configuration $\{\bm{\theta}, \bm{x}_{cog}, \bm{x}_{tool}, \bm{s}_{tool}\}\rightarrow\{\bm{\theta}, \bm{x}_{cog}, \bm{x}_{tool}, \bm{s}_{tool}\}$.
  Compared to the experiments in PR2, the addition of information on the center of gravity and the visual position of the tool enables a more diverse description of the loss function.
  By changing the weight and length of the tools, the tool information is embedded in the parametric bias as before, and this information can be correctly recognized from the changes in the center of gravity and the tool-tip position in the visual field.
  In particular, it was found that the more information available, the faster the online update of parametric bias converges, and the less information available, the longer it takes to recognize the tools.
  By controlling not only the tool-tip position but also the center-of-gravity position, it becomes possible for even a robot with low rigidity to perform stable tool manipulation.
  Also, by fine-tuning the simulation model with the actual robot data, more accurate tool manipulation control becomes possible.
}%
{%
  これまでに得られた実験結果をまとめる.
  まずPR2による道具先端操作実験では, $\bm{\theta}\rightarrow\bm{x}_{tool}$という非常に単純なネットワーク構成について扱った.
  $\bm{\theta}\rightarrow\bm{x}_{tool}$は計算可能であるが, その逆は計算できないということが自動的に検出され, モデル構築が行われた.
  多様な把持状態においてデータを取得することで, それら把持状態変化の情報がParametric Biasに埋め込まれ, これをオンライン更新することで, 常に現在の把持状態を知り続けることができる.
  また, ネットワークのforwarding, またはbackpropagationとgradient descentに基づいて道具先端位置の推定と制御が可能であり, これらは把持状態を正しく認識することでより正確になる.
  オンライン更新はParametric Biasのみに対して行うことで高い汎化能力を継続できるが, ネットワークの重み$\bm{W}$を更新した場合は汎化能力を失いやすい.
  加えて, 実機実験では柔軟な道具を扱ったが, 同様にParametric Biasのオンライン更新・道具先端制御が可能であり, その性能が示された.

  次に筋骨格ヒューマノイドMusashiの動作制御実験では, $(\bm{\theta}, \bm{f}, \bm{l})\rightarrow(\bm{\theta}, \bm{f}, \bm{l})$というより複雑なネットワーク構成について扱った.
  筋骨格系における$\{\bm{\theta}, \bm{f}, \bm{l}\}$の間には非常に複雑な関係があり, 一般的に$\{\bm{\theta}, \bm{f}\}$から$\bm{l}$が, $\{\bm{f}, \bm{l}\}$から$\bm{\theta}$が, $\{\bm{\theta}, \bm{l}\}$から$\bm{f}$が計算できる.
  これらを一つのネットワークで表現することで, 状態推定や制御, シミュレーションが可能となった.
  幾何モデルから得られたデータを用いてネットワークを初期化し, それを実機データを使ってオンラインで更新する.
  前述の実験とは異なりParametric Biasを含まないため$\bm{W}$を直接更新する例であるが, ランダムな関節角度と筋張力指令によってデータを収集することで, 過適合を避けることができる.
  実機センサデータからネットワークを更新することで, 状態推定や制御, シミュレーションの精度が向上した.
  また, シミュレーションについては損失関数の定義次第で多様な状況をシミュレートでき, 提案手法の汎用性が示された.

  最後に, 低剛性ヒューマノイドKXRの全身道具操作実験では, $\{\bm{\theta}, \bm{x}_{cog}, \bm{x}_{tool}, \bm{s}_{tool}\}\rightarrow\{\bm{\theta}, \bm{x}_{cog}, \bm{x}_{tool}, \bm{s}_{tool}\}$という非常に複雑なネットワーク構成について扱った.
  PR2における実験に加え, 足裏重心位置や視覚上における道具位置情報が加わることで, より多様な損失関数記述が可能になる.
  道具の重さや長さを変化させることで, これまで同様道具情報がParametric Biasに埋め込まれ, これを重心変化や視覚上に道具先端位置変化から正しく認識することができる.
  特に, Parametric Biasのオンライン更新の際には, 得られる情報が多ければ多いほどオンライン更新の収束が早く, 情報が少ないと認識に非常に時間がかかることがわかった.
  また, 道具先端位置だけでなく重心位置までも制御することで, 低剛性なロボットでも安定して道具操作を行うことが可能になる.
  加えて実機実験では, シミュレーションで得られたモデルを実機データによってfine-tuningすることで, より正確な道具操作制御が可能となる.
}%

\subsection{Limitations}

\switchlanguage%
{%
  Based on these experimental results, we discuss the limitations and future prospects of this study.

  First, data collection was generally performed by random action or human teaching using GUI.
  On the other hand, the concept of reinforcement learning is also effective in autonomously collecting valid data.
  We believe that the combination of this research with reinforcement learning will lead to a more practical system capable of generating complex motions.

  Second, regarding catastrophic forgetting, updating only the low-dimensional latent space poses no problem when using parametric bias.
  However, updating the overall weight $\bm{W}$ may lead to issues with catastrophic forgetting.
  Techniques such as Elastic Weight Consolidation \cite{kirkpatrick2017ewc} have been developed, and their incorporation should be considered in the future.

  Third, regarding increasing sensor numbers, the current configuration results in an exponential increase in the number of masks based on sensor quantity, limiting the ability to infinitely add sensors.
  As the complexity of the body increases, the increase in sensor numbers becomes unavoidable, necessitating the development of more efficient learning methods.

  Fourth, regarding anomaly detection, it is not possible to differentiate between anomalies and dynamic environmental changes.
  An anomaly is an unpredicted change, and to avoid categorizing an event as an anomaly, it is necessary to include observable sensor values for that event in the network input and output.
  However, the current setup only includes primitive sensors, and the incorporation of depth sensors, audio information, etc., will be necessary in the future.

  Finally, to elevate the system to a more practical level, further application to diverse bodies, environments, tasks, and experiments is essential.
  Also, contributions to cognitive science and a deeper understanding of the relationship with the human brain are areas of interest for future research.
}%
{%
  これらの実験結果を受け, 本研究の限界と今後の展望について述べる.

  まず, データ収集方法についてである.
  本研究はデータ収集を基本的にランダム動作またはGUIによる人間の操作によって行った.
  一方で, 自律的に有効なデータを集める強化学習の考え方ももちろん有効であり, 今後の適用はより複雑な動作生成への鍵となろう.

  次に, 破壊的忘却についてである.
  parametric biasを用いる場合は低次元の潜在空間のみ更新されるため問題ないが, 全体の重み$\bm{W}$を更新する場合は破壊的忘却が問題となる.
  Elastic Weight Consolidation \cite{kirkpatrick2017ewc}等の手法が開発されており, 今後それらの導入も視野に入れたい.

  次に, センサ数増加の問題についてである.
  現状の構成では, センサ数に応じて指数関数的にマスク数が増えるため, 無限にセンサ数を増やせるわけではない.
  今後より複雑な身体を扱うに従い, センサ数の増加は避けられないため, より効率的な学習方法が求められる.

  次に, 異常検知についてである.
  現状, 異常と動的な環境変化を区別することはできない.
  異常とは予測できていない変化であり, ある事象を異常として判断したくなければ, その事象を観測可能なセンサ値をネットワーク入出力に加える必要がある.
  一方で, 現状はprimitiveなセンサのみであり, 深度センサや音声情報などは今後取り入れていく必要がある.

  最後に, より実用的なシステムへと昇華させるためには, さらなる多様な身体や環境, タスクへの適用, 実験が必要であると考えている.
  また今後, 認知科学等への貢献も行っていきたく, 実際の人間や動物の脳との関係についても研究を進めたい.
}%

\section{CONCLUSION} \label{sec:conclusion}
\switchlanguage%
{%
  In this study, we have developed a method for robot control, state estimation, anomaly detection, simulation, and environmental adaptation by learning a body schema that describes the correlations between sensors and actuators of the robot's body, tools, and environment.
  By using a mask variable as input to the network, correlations between sensory and control input data can be described in the network.
  By using parametric bias, it is possible to incorporate the implicit changes in the correlation between body, tool, and environment into the model.
  By using the iterative backpropagation and gradient descent method, control, state estimation, anomaly detection, and simulation can be performed based on this single body schema.
  By updating the network weight and parametric bias, we can cope with changes in the grasping state of the object, changes in the characteristics of the tool, aging of the body, etc.
  With this method, we have succeeded in learning adaptive tool-tip control considering the changes in grasping state of an axis-driven robot, learning joint-muscle mapping for a musculoskeletal humanoid, and full-body tool manipulation considering tool changes for a low-rigidity plastic-made humanoid.
}%
{%
  本研究では, ロボットの身体-道具-環境に関する感覚と運動の相互関係を記述する身体図式を学習することで, ロボットの制御や状態推定, 異常検知, シミュレーション, 環境適応等まで行う手法を開発した.
  ネットワークの入力にマスク変数を用いることで, 感覚と運動のそれぞれの値に関する相関関係をネットワーク内に記述することができる.
  また, Parametric Biasを用いることで, 身体-道具-環境の相関関係変化モデル内に取り込むことが可能である.
  繰り返しの誤差逆伝播と勾配法を用いることで, この身体図式をもとに制御, 状態推定, 異常検知, シミュレーションを行うことが可能となった.
  ネットワーク重みやParametric Biasを更新することで, 物体の把持状態変化や道具特性の変化, 身体の経年劣化等にも対応することができるようになった.
  本手法により, 軸駆動型ロボットの把持状態を考慮した適応的道具先端操作学習, 筋骨格ヒューマノイドの逐次的関節-筋空間マッピング学習, 低剛性樹脂製ヒューマノイドの道具変化を考慮した全身道具操作に成功した.
}%

{
  %\footnotesize
  %\small
  %\bibliographystyle{junsrt}
  \bibliographystyle{IEEEtran}
  \bibliography{main}
}

\end{document}